\documentclass{article}

\PassOptionsToPackage{numbers, compress}{natbib}

\usepackage[preprint]{neurips_2021}
\usepackage{hyperref}

\usepackage{mathtools} 
\usepackage{booktabs} 
\usepackage{tikz} 
\usepackage{wrapfig, blindtext}

\newcommand{\mbf}[1]{\mathbf{#1}}
\newcommand{\cut}[1]{}

\DeclareMathOperator*{\argmax}{arg\,max}
\DeclareMathAlphabet{\pazocal}{OMS}{zplm}{m}{n}
\usepackage{subcaption}
\usepackage{amssymb} 
\usepackage{amsmath} 
\usepackage{multirow}

\usepackage[noend]{algpseudocode}
\usepackage{algorithm}
\usepackage{subcaption}

\title{Impact of Parameter Sparsity on Stochastic Gradient MCMC Methods for Bayesian Deep Learning}

\author{
  Meet P. Vadera \\
  University of Massachusetts Amherst\\
  \texttt{mvadera@cs.umass.edu} \\
  \And
  Adam D. Cobb \\
  SRI International\\
  \texttt{adam.cobb@sri.com} \\
   \AND
  Brian Jalaian \\
  US Army Research Laboratory\\
  \texttt{brian.a.jalaian.civ@mail.mil} \\
   \And
   Benjamin M. Marlin \\
   University of Massachusetts Amherst \\
   \texttt{marlin@cs.umass.edu}\\
}

\begin{document}
\maketitle

\begin{abstract}
  Bayesian methods hold significant promise for improving the uncertainty quantification ability and robustness of deep neural network models. Recent research has seen the investigation of a number of approximate Bayesian inference methods for deep neural networks, building on both the variational Bayesian and Markov chain Monte Carlo (MCMC) frameworks. A fundamental issue with MCMC methods is that the improvements they enable are obtained at the expense of increased computation time and model storage costs. In this paper, we investigate the potential of sparse network structures to flexibly trade-off model storage costs and inference run time against predictive performance and uncertainty quantification ability. We use stochastic gradient MCMC methods as the core Bayesian inference method and consider a variety of approaches for selecting sparse network structures. Surprisingly, our results show that certain classes of randomly selected substructures can perform as well as substructures derived from state-of-the-art iterative pruning methods while drastically reducing model training times.
  
\end{abstract}
\section{Introduction}
Modern deep neural network architectures are known to exhibit poor uncertainty calibration \citep{guo2017calibration} and can fail to be robust to out-of-distribution examples and adversarial manipulation \citep{Goodfellow2015ICLR}. Bayesian methods hold significant promise for improving the uncertainty quantification performance and robustness of deep neural network models due to their ability to reflect model uncertainty in posterior predictions as well as their ability to decompose uncertainty in a way that is inaccessible to point-estimated models \citep{Depeweg2017DecompositionOU}. 

Due to these potential advantages, there has been significant recent interest in the application of both the variational Bayesian  \citep{jordan1999introduction} and Markov chain Monte Carlo (MCMC) \citep{smith1993bayesian} approximate inference frameworks to the problem of Bayesian deep learning (BDL). A fundamental issue with such methods is that the improvements they enable are obtained at the expense of increased computation time and model storage requirements. These issues are particularly acute for MCMC-based methods as a Monte Carlo approximation using $S$ samples from the model posterior requires $S$ times more storage and $S$ times more computation to produce predictions than a corresponding base architecture. The efficiency of MCMC sampling methods also tends to decrease as model sizes increase. Prior work has addressed the deployment-time storage and computational cost of MCMC-based deep learning using posterior distillation methods \citep{balan2015bayesian,Vadera2020GeneralizedBP}, while work on subspace inference methods \citep{Izmailov2019SubspaceIF} has addressed the problem of sampling efficiency by sampling in significantly lower dimensional spaces. 

In this paper, we consider a different approach to obtaining more compact approximations to model posteriors based on performing inference within sparse substructures contained within larger networks. Prior research on weight-level sparsity in deep neural networks suggests that deep network architectures often contain much more compact sub-networks that accounts for a significant majority of their predictive accuracy \citep{Han2015LearningBW,Frankle2018TheLT}. Inspired by these observations, we study the performance trade-offs enabled by different methods for selecting sparse structures within large networks. We report trade-offs between sparsity level, predictive accuracy, predictive log likelihood, calibration error, training time, test time computational efficiency, and mixing efficiency.

We consider the use of state-of-the-art iterative pruning approaches to derive sparse network structures as a pre-processing step. We show that while these methods can produce high-quality and highly sparse sub-structures, their training time can be prohibitively high. We subsequently show the surprising result that certain classes of randomly selected sub-structured can perform as well over a range of target sparsity rates.  

\section{Background \& Related Work}
\label{backgroun}
In this section, we review background and related work on Bayesian supervised deep learning, approximate inference, and pruning of neural network models.

\subsection{Bayesian Deep Supervised Learning}
We begin by introducing the notation used in this paper. Let $\mathcal{D}$ be a data set containing of a collection of labeled pairs $\{(\mathbf{x}_i,y_i)\}_{i=1}^N$ where $\mathbf{x}_i\in\mathbb{R}^D$ are feature vectors and $y_i\in\mathcal{Y}$ are prediction target values. A Bayesian deep supervised learning model requires the specification of two model components \citep{Neal:1996:BLN:525544}. The first component is a conditional probability model of the form $p(y|\mathbf{x}, \theta)$ parameterized using a deep neural network with parameter vector $\theta\in\mathbb{R}^K$. The second component is a prior distribution over the neural network model parameters $p(\theta|\gamma)$. $\gamma$ are the hyper-parameters of the prior. An isotropic Gaussian prior is often used in Bayesian deep learning applications.

The primary Bayesian supervised learning inference problem is the computation of the posterior predictive distribution $p(y| \mathbf{x}, \mathcal{D},\gamma)$, which is defined  in terms of the parameter posterior $p(\theta|\mathcal{D},\gamma)$, as shown in Equations \eqref{eq:posterior_predictive} and \eqref{eq:posterior}.
\begin{align}
\label{eq:posterior_predictive}
&p(y| \mathbf{x}, \mathcal{D},\gamma) = 
\mathbb{E}_{p(\theta|\mathcal{D},\gamma)}[p(y|\mathbf{x}, \theta)]\\
\label{eq:posterior}
    &p(\theta|\mathcal{D},\gamma) = 
      \frac{p(\theta|\gamma)\cdot \prod_{i=1}^N p(y_i|\mathbf{x}_i, \theta)}
    {\int p(\theta'|\gamma)\cdot \prod_{i=1}^N p(y_i|\mathbf{x}_i, \theta') d\theta'}
\end{align}
In the case of Bayesian supervised learning using common deep neural network architectures, the integrals required in Equations \eqref{eq:posterior_predictive} and \eqref{eq:posterior} are intractable and require approximation. The two dominant approaches to approximate Bayesian inference are variational Bayesian (VB) methods and Markov chain Monte Carlo (MCMC) methods. VB methods introduce a computationally tractable approximate posterior distribution $q(\theta|\mathcal{D},\gamma,\phi)$ and aim to minimize the Kullback–Leibler divergence from $q(\theta|\mathcal{D},\gamma,\phi)$ to  $p(\theta|\mathcal{D},\gamma)$ with respect to the introduced variational parameters $\phi$ \citep{jordan1999introduction, jaakkola2000bayesian, graves2011practical, blundell2015weight}. Recent methods include MC-dropout, a stochastic VB method that is particularly easy to implement \citet{gal2016dropout}.


In this work, we focus on the MCMC family of approximate inference methods, which, relative to VB methods, trade off longer running times and expanded storage requirements for unbiased representations of the model posterior \citep{casella1992explaining,smith1993bayesian,chib1995understanding,duane1987hybrid,neal2003slice,girolami2011riemann}. MCMC methods simulate a Markov chain with equilibrium distribution equal to the true parameter posterior $p(\theta|\mathcal{D},\gamma)$ in order to draw a set of $S$ posterior samples $\theta^{(1)},...,\theta^{(S)}$. These samples can be used to approximate posterior statistics including the posterior predictive distribution  $\hat{p}(y| \mathbf{x}, \mathcal{D},\gamma)
=\frac{1}{S}\sum_{s=1}^S p(y|\mathbf{x}, \theta_s)$.

However, the classical MCMC methods referenced above are typically not computationally efficient enough to be used in Bayesian deep learning applications. In this work, we focus on the stochastic gradient MCMC family of approaches, which leverage a stochastic gradient approximation to the unnormalized model posterior to accelerate inference computations relative to methods that need to compute over the complete data set \citep{welling2011bayesian, Chen2014StochasticGH, Zhang2020Cyclical}. We select stochastic gradient Hamiltonian Monte Carlo (SGHMC) as a representative of this family of methods as it has been shown to outperform more basic stochastic gradient approaches with little additional computational cost \citep{Chen2014StochasticGH}.

However, a number of issues remain in terms of practical use of MCMC methods for larger-scale deep models. First, storing the posterior ensemble requires storage proportional to $S\cdot K$. Second, computing the posterior predictive distribution is $S$ times slower than using a single point estimated model due to the Monte Carlo approximation exhibited earlier.  Third, the underlying Markov chains can mix poorly in large parameter spaces. We note that the linear increase in storage and prediction run time can prohibit the application of Bayesian deep learning in embedded, mobile and IoT contexts, where there is high sensitivity to prediction computation time and latency. For example, benchmarking results for a range of detection models on the Jetson Nano edge device show frame rates between 5 and 40 frames per second.\footnote{\url{https://developer.nvidia.com/embedded/jetson-nano-dl-inference-benchmarks}} When using a 100 element Bayesian ensemble of the same models, we would expect the frame rates to scale down linearly to between 0.05 and 0.4 frames per second (between 20 and 2.5 seconds per frame), far slower than would be acceptable for real-time applications. 

The recent literature has suggested several approaches to dealing with these issues. The subspace inference method developed by \citet{Izmailov2019SubspaceIF} identifies a low-dimensional subspace of the full parameter space, samples within this subspace, and finally projects the samples back to the full parameter space for evaluation. This can result in a compressed representation of the posterior ensemble and enables improved sampling efficiency, but does not address the issue of the complexity of computing posterior statistics due to the required projection back to full parameter space. Posterior distillation methods instead attempt to decrease deployment time computational complexity and storage cost by distilling the computation of posterior statistics into a single deep neural network of reduced size relative to a Monte Carlo posterior ensemble \citep{balan2015bayesian,Vadera2020GeneralizedBP}. However, distillation methods are reliant on a base approximate inference method to obtain samples from the true posterior and thus don't address the problem of efficiency during sampling. Furthermore, as \citet{Vadera2020GeneralizedBP} showed, identifying the correct student model architecture can also pose a challenge for posterior distillation. The approach that we consider is based on identifying sparse neural network sub-structures followed by performing SGHMC-based inference within these structures. This approach simultaneously addresses the need for compact posterior representations, decreased deployment time computational complexity when computing posterior statistics, and has the potential to improve sampling efficiency. Next, we briefly discuss methods for identifying sparse neural network substructures.

\subsection{Sparsity and Neural Networks}

Identifying sparse neural network structures has an extended history in the machine learning community \citep{LeCun1989OptimalBD, Hassibi1993OptimalBS} and a wide variety of methods for learning sparse neural network models have been proposed in the literature. The two main categories of approaches are methods that attempt to infer sparsity at the node level and thereby prune out complete neurons and methods that attempt to infer sparsity at the weight level and thereby prune connections between individual nodes. For example \citet{Hassibi1993OptimalBS} use second-order derivatives of the objective function to guide the pruning of network connections, while $\ell_1$ sparsity methods related to the LASSO have been applied to prune networks both at the weight level and the node level \citep{Zhang2018LEARNINGSS,alvarez2016learning,wen2016learning,He2017ChannelPF}. In a closely related approach \citet{Louizos2017BayesianCF} use hierarchical sparsity inducing priors within VB methods to perform node-level pruning.

One of the key observations in the recent literature is that predictive performance appears to be better maintained by iterative pruning approaches that alternate multiple rounds of learning and pruning as compared to methods that use a single pruning step \citep{Hassibi1993OptimalBS}. Indeed, \citet{Frankle2018TheLT} use a particular form of iterative pruning that rewinds weights to their initial random values after each pruning iteration to show that it is possible to identify sparse sub-structures that can be trained from matched random initializations while still achieving strong predictive performance. They further show that performance degrades when identified sub-structures are paired with newly randomized initial parameters, indicating increased sensitivity to initial parameters when attempting to optimize parameters over sparse sub-structures. In this work, we investigate weight-sparse network structures derived using iterative pruning as well as structures sampled at random. 

Lastly, there's an additional important line of work that looks at implementing sparse neural networks of different architectures to leverage sparse substructures in models and achieve inference time speedups \citep{Park2016HolisticSF,park2016faster,li2017enabling}. \citet{park2016faster} present a pruning algorithm along with an implementation of efficient sparse-dense matrix multiplication that helps them achieve speedups in the range  $\approx (3.1-7.3)\times$ for deep convolutional models with approximately 90\% weight sparsity on Intel Atom, Xeon, and Xeon Phi processors. \citet{Park2016HolisticSF} provide a holistic approach that looks at training by sparsifying all layers simultaneously while controlling performance metrics. They present a highly optimized sparse CNN implementation which provides a speedup of $\approx 3.4\times$ on Intel Atom and a speedup of $\approx 7.1\times$ on Intel Xeon processor while using the AlexNet model \citep{krizhevsky2012imagenet}, also with approximately weight 90\% sparsity. 

\section{Methods}

In this section, we describe the methods that we apply to study the properties of stochastic gradient MCMC  applied to substructures of sparse deep neural networks. We consider a two-step process. Given an initial model architecture, we first apply a method to select a sparse substructure from the initial dense model. We select sparse substructures at the connection level using weight-level sparsity masks. We then study the properties of Bayesian inference applied within the selected substructures. In this section, we first describe methods for selecting sparse network structures. We then provide details for the stochastic gradient MCMC method that we apply. 

\subsection{Selecting Sparse Substructures}
\label{sec:methods:substructs}
We let $\theta_l\in\mathbb{R}^{K_l}$ be the parameter vector for layer $l$ and $K_l$ be the number of parameters in layer $l$. We define $\mbf{m} \in \{0,1\}^K$ to be a weight-level sparsity mask and $\mbf{m}_l \in \{0,1\}^{K_l}$ be the sparsity mask for layer $l$. For defining parameters and weight-level sparsity masks at iteration $i$, we use the notation $\theta^i$ and $\mbf{m}^i$ respectively. Given a sparsity mask $\mbf{m}$ and a parameter vector $\theta$, the masked parameters are given by the Hadamard product $\tilde{\theta}= \mbf{m}\odot\theta$. Below we briefly describe several approaches to selecting sparse sub-network structures. We emphasize that the contribution of this work is not to propose new methods for deriving sparse sub-networks, but rather to evaluate the impact of sparse sub-network structures on MCMC-based inference methods, a question that has not been addressed in prior research.

\textbf{Iterative Pruning (IP):} The first approach that we consider for selecting a sparsity mask is based on the iterative pruning method of \citet{Han2015LearningBW}. The iterative pruning approach can use any standard gradient-based optimization method as the base learning method. The algorithm begins with a randomly selected parameter vector $\theta^0\in\mathbb{R}^K$ with all weights active ($\mbf{m}^0 = 1$). The algorithm proceeds over $T$ pruning iterations. In each iteration $i$, a specified number of epochs $\epsilon$ of the base learning method are applied yielding a final parameter vector $\theta^i$ for that iteration. At the end of the iteration, a fixed fraction $\pi$ of the active weights are pruned resulting in a new sparsity mask $\mbf{m}^{i+1}$. The active weights with the smallest magnitudes are selected for pruning on each iteration. The initial weight vector for iteration $i+1$ is then given by  $\theta^i \odot \mbf{m}^{i+1}$. This method greedily produces a nested sequence of increasingly sparser network substructures encoded by the sparsity masks $\mbf{m}^i$ along with corresponding locally optimal parameters $\theta^i$.

\textbf{Iterative Pruning with Rewinding (IPR):} \citet{Frankle2018TheLT} proposed an iterative pruning method that is very similar to the method of \citet{Han2015LearningBW}, which we will refer to as Iterative Pruning with Rewinding. This approach differs from basic iterative pruning in that following each pruning iteration, the weights used to initialize the next iteration of the algorithm are formed by combining the updated mask $\mbf{m}^{i+1}$ with the original random weight vector $\theta^0$ instead of the weight vector obtained at the end of iteration $i$. Effectively, the active weights are rewound to their initial values at the start of each round of iterative pruning. We will again define $\theta^i$ to be the value of the weights at the end of iteration $i$. 
This method also greedily produces a nested sequence of increasingly sparse network structures encoded by the sparsity masks $\mbf{m}^i$ along with corresponding locally optimal point-estimated parameters $\theta^i$. However, as \citet{Frankle2018TheLT} show, the pairs  $(\theta^0, \mbf{m}^i)$ appear to have somewhat special properties with respect to optimization-based learnability. We investigate whether the properties of the sub-structures and initial parameters identified by iterative pruning with rewinding differ from those identified by iterative pruning when sparse sub-structures are used in the context of MCMC-based Bayesian deep learning.

\textbf{{Random Layer-wise Masking (RLM):}} A question that is unexplored in the both \citet{Han2015LearningBW} and \citet{Frankle2018TheLT} from the optimization perspective  is the importance of the exact sparse structure obtained by iterative pruning relative to the sparsity level per network layer. We study a Layer-wise Random Masking approach to answer this question in the case of Bayesian deep learning. This approach requires the specification of a desired sparsity rate $\pi_l$ for each layer $l$ of the network. Given the desired sparsity rate $\pi_l$, we uniformly sample a random sparsity mask $\mbf{m}_l$ for layer $l$ from the space of all binary vectors with this sparsity rate. These layer-wise random masks are then assembled into a sparsity mask $\mbf{m}$ for the complete network. We explore three approaches to select the desired sparsity levels $\pi_l$. The first approach runs iterative pruning for $T$ iterations, obtains the final sparsity mask, and computes the corresponding sparsity rate $\pi_l$ for each layer. We denote this approach by RLM(IP). Iterative pruning with rewinding can be used in the same way, obtaining an approach we denote by RLM(IPR). These two approaches allow us to compare the optimized sparse network structures obtained using iterative pruning to networks where the sparsity rates per layer have been optimized while the specific sparsity structures are randomly selected. The final variant of the random layer-wise approach simply sets the same fixed sparsity rate per level $\pi_l=\pi$. This approach allows us to investigate how critical preserving the per-layer sparsity rates are. We refer to this approach as RLM(F).

\noindent\textbf{{Random Global Masking (RGM):}} Random global making is the most basic mask generation approach that we consider. We specify a fixed sparsity rate $\pi$ and select a random mask $\mbf{m}\in\{0,1\}^K$ from the space of all binary vectors yielding this sparsity level. This approach allows us to study how critical the distribution of sparsity over network levels is compared to the overall level of sparsity. 
As we conclude this subsection, we emphasize that RLM(F) and RGM do not require us to run any form of iterative pruning, resulting in significant training time and resource savings when compared to the other described methods.

\subsection{SGHMC in Sparse Structures}
\label{sec:methods:inference}

Given a sparse neural network structure represented by a weight-level sparsity mask $\mbf{m}$, we perform stochastic gradient Hamiltonian Monte Carlo (SGHMC) within the specified substructure. SGHMC only requires the computability of an unnormalized version of the log parameter posterior density:
$U(\theta) =  -\log p(\theta|\gamma) -\sum_{i=1}^N p(y_i|\mathbf{x}_i, \theta)$.
Like all HMC methods, SGHMC is based on sampling in an extended space that includes the model parameters $\theta\in\mathbb{R}^K$ as position variables as well as an auxiliary set of momentum variables $\mbf{r}\in\mathbb{R}^K$. This sampling process simulates discretized Hamiltonian dynamics. SGHMC methods compute a stochastic approximation the gradient $\nabla U(\theta)$  of the unnormalized energy function $U(\theta)$ using a mini-batch of data $\mathcal{D}'_{tr}\subset \mathcal{D}_{tr}$ of size $B$. When performing Bayesian inference over sparse substructures, the stochastic gradient $\nabla \tilde{U}(\theta)$ only needs to be computed over the selected substructure and doing so has the potential to significantly accelerate inference time. For simplicity of implementation, we use a less computationally efficient masking approach during training that allows existing neural network model implementations to be used. 

The complete SGHMC algorithm that we use for sampling within substructures is defined in Algorithm \ref{sghmc_alg} in Appendix \ref{app:sparse_sghmc}. In this algorithm, $\theta^{(0)}$ is an initial parameter vector with sparsity structure matching that specified by the sparsity mask $\mbf{m}$. $\alpha_s$ is the step size used on sampling iteration $s$. $\eta$ is a friction term introduced to counteract the additional noise introduced by using stochastic gradients. $S$ is the number of desired sampled. 
Algorithm \ref{sghmc_alg} adds a masking step to the basic SGHMC algorithm used by \citet{Zhang2020Cyclical}. The masking step projects sampled parameters back into the required substructure on each sampling iteration. This algorithm correctly samples from the posterior distribution of the weights in the selected substructure due to the fact that the updates for both $\mbf{r}$ and $\theta^{(s)}$ do not mix information across their dimensions.
For large models, running SGHMC from a randomly chosen $\theta^{(0)}$ can result in the need for very long burn-in times. To help mitigate this issue, we always initialize $\theta^{(0)}$ using the output of an SGD-based optimization algorithm applied within the substructure corresponding to $\mbf{m}$. This optimization step itself requires an initial setting of the model parameters. We initialize at random for random masks. For iterative pruning-based methods, we use the final parameter vector returned along with the sparsity mask.

\section{Experiments and Results}
\label{sec:experiments}
\label{sec:experiments:protocols}

In this section we present experiments and results. Each of the experiments involves the composition of several components including a base neural network architecture, a data set, a method for selecting a sparse substructure (see Section \ref{sec:methods:substructs}), and an optimization or Bayesian inference approach (see Section \ref{sec:methods:inference}). We focus on classification as a task and use three model/data set combinations. We describe these combinations below. We use an isotropic Gaussian prior for all models. 

To evaluate the effect of sparsity on Bayesian inference in fully connected architectures, we use the Fashion MNIST data set (FMNIST) \citep{Xiao2017FashionMNISTAN} with a two-hidden layer MLP with 200 hidden units per hidden layer (MLP200). The FMNIST data set has 10 classes, 60K training instances and 10K test instances. 
For a larger-scale model, we pair the 20-layer residual network model (ResNet20) used in \citet{Frankle2018TheLT} with the CIFAR10 data set \citep{krizhevsky2009learning}. CIFAR10 has 10 classes, 50K training instances and 10K test instances. For a more challenging data set, we pair the ResNet20 model with the CIFAR100 data set \citep{krizhevsky2009learning}. CIFAR100 has 100 classes, 50K training instances and 10K test instances. All data sets are freely available.
\begin{wraptable}{r}{0.55\textwidth}
\small
\center
\caption{Total no. of parameters by sparsity rate ($\downarrow$) and wall-clock time per image during testing ($\downarrow$).}
\resizebox{0.5\textwidth}{!}{
\begin{tabular}{lllrr}
\toprule
Dataset & Method & Sparsity & \# Params & \begin{tabular}[c]{@{}c@{}}Inference\\ speedup\end{tabular} \\ 
\midrule
FMNIST  & Full-SGHMC & 0\%  & 9960500  & $1.0\times$ \\
FMNIST  & IP-SGHMC   & 83\% & 1695300  & $1.1\times$     \\
FMNIST  & IP-SGHMC   & 89\% & 1085700  & $1.6\times$ \\
FMNIST  & IP-SGHMC   & 95\% & 468150   & $16.0\times$ \\
FMNIST  & Full-OPT   & 0\%  & 199210   & $50.0\times$ \\ \hline
CIFAR10 & Full-SGHMC & 0\%  & 13623700 & -     \\
CIFAR10 & IP-SGHMC   & 83\% & 2318800  & -     \\
CIFAR10 & IP-SGHMC   & 89\% & 1485000  & -     \\
CIFAR10 & IP-SGHMC   & 95\% & 640350   & -     \\
CIFAR10 & Full-OPT   & 0\%  & 272474   & -     \\
\bottomrule
\end{tabular}
}
\label{tab:exp2b}
\end{wraptable}

For evaluating performance on the classification task, we use the metrics of accuracy ($\uparrow$), negative log likelihood (NLL, $\downarrow$), and expected calibration error (ECE, $\downarrow$). For assessing the efficiency of MCMC samplers, we use the metrics of effective sample size (ESS, $\uparrow$), and auto-correlation coefficient (ACF $\downarrow$). We provide detailed definitions of these metrics in Appendix \ref{app:methods:metrics}. Additional implementation details and training procedures for our experiments are presented in Appendix \ref{app:implementation_details}. We now turn to describing the experiments and results.


\begin{figure*}[t]
\center
\includegraphics[width=\textwidth]{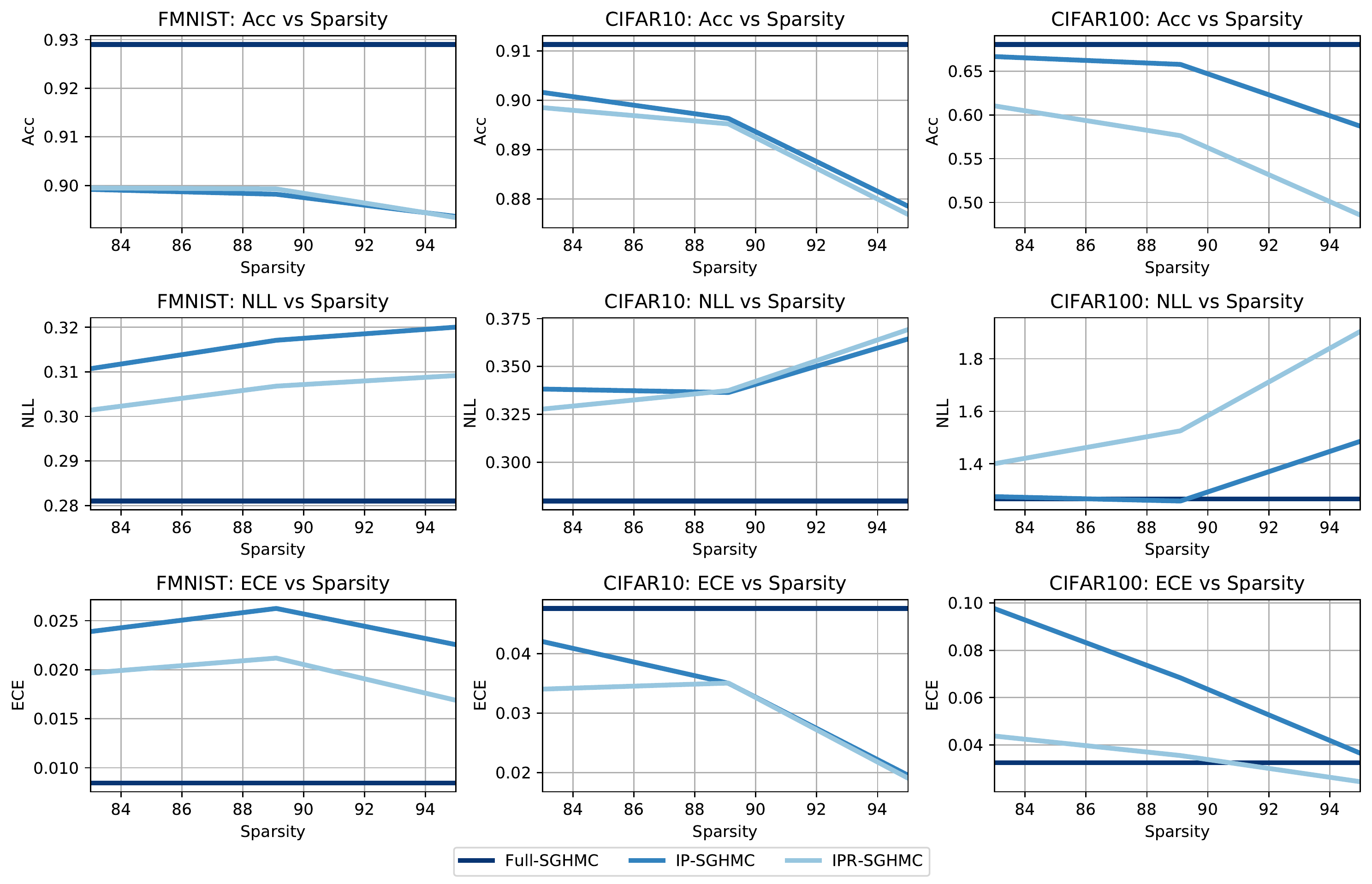}
\caption{Performance of SGHMC 
applied to optimized sparse substructures compared to SGHMC applied to full models.}
\label{fig:exp2}
\end{figure*}

\textbf{Experiment 1: How does SGHMC perform when applied to optimized sparse substructures?} In this experiment, we evaluate the performance of SGHMC applied to substructures selected using the iterative pruning (IP-SGHMC) and iterative pruning with re-winding (IPR-SGHMC) methods. We compare to the baseline of SGHMC applied to the full, dense model (Full-SGHMC). The results are presented in Figure \ref{fig:exp2}. As we can see, the drop in accuracy at 83\% sparsity is less than 3\% on FMNIST and CIFAR10 for both IP-SGHMC an IPR-SGHMC. On CIFAR100, IP-SGHMC performs well at 83\% sparsity while IPR-SGHMC shows a larger drop. As expected, both methods show decreased accuracy as sparsity increases on all data sets. However, the drop in accuracy is minimal (less than 4\%) on both FMNIST and CIFAR10 even at 95\% sparsity. The degradation in accuracy is more pronounced on the more challenging CIFAR100 data set. We can also see that the negative log likelihood increases for both methods on both data sets as a function of sparsity. However, there is no clear advantage for either method. Finally, we can see that the expected calibration errors are very low on both FMNIST and CIFAR10 for both methods in terms of absolute level, while on CIFAR100 IP-SGHMC appears to exhibit poor calibration at lower sparsity levels.

To provide context for the trade-off between storage and performance, in Table \ref{tab:exp2b} we show the total number of parameters that require storage under Full-SGHMC and IP-SGHMC on FMNIST and CIFAR10. The parameter counts for IPR-SGHMC are identical to those of IP-SGHMC and the parameters required for CIFAR100 are nearly identical to those required for CIFAR10 for all methods. We also include the parameter count for the base models (Full-OPT) for reference. Importantly, for the SGHMC methods, we report the total number of parameters in the complete sparse ensemble. As we can see, even at the lowest level of sparsity tested (83\%), the savings in storage for the sparse posterior ensembles is highly significant compared to the dense posterior ensembles. At 95\% sparsity, the sparse posterior ensembles have roughly twice the storage cost of the base model (Full-OPT). The trade-off between storage and performance appears to be quite favorable for both FMNIST and CIFAR10 at the lower sparsity levels relative to the full posterior ensemble. 

We also provide a prediction inference time speedup comparison relative to the dense ensemble (Full-SGHMC) for the MLP200/FMNIST model in Table \ref{tab:exp2b}. For computing the speedups, we implemented the MLP200 model using the GNU Scientific Library's (GSL) sparse matrix multiplication functions \citep{kernighan2006c, galassi2003gnu} and ran experiments on an Apple Macbook Pro (8 GB Memory, M1 processor). As we can see, at 89\% sparsity we achieve an $\approx 1.6\times$ speedup, while at 95\% sparsity we achieve $\approx 16\times$ speedup. These are highly significant improvements in prediction inference time relative to the modest drop in prediction performance on FMNIST.\footnote{We note that we do not include speedup results for convolutional models, but prior research shows these speedups to be in the range of $3-7\times$  as noted previously \citet{park2016faster, li2017enabling}.}  
We present additional details for this experiment in Appendix \ref{app:implementation_details}, and an expanded set of results in Appendix \ref{app:additional_results} (Table \ref{tab:exp1_timing}).





\begin{figure*}[t]
\center
\includegraphics[width=\textwidth]{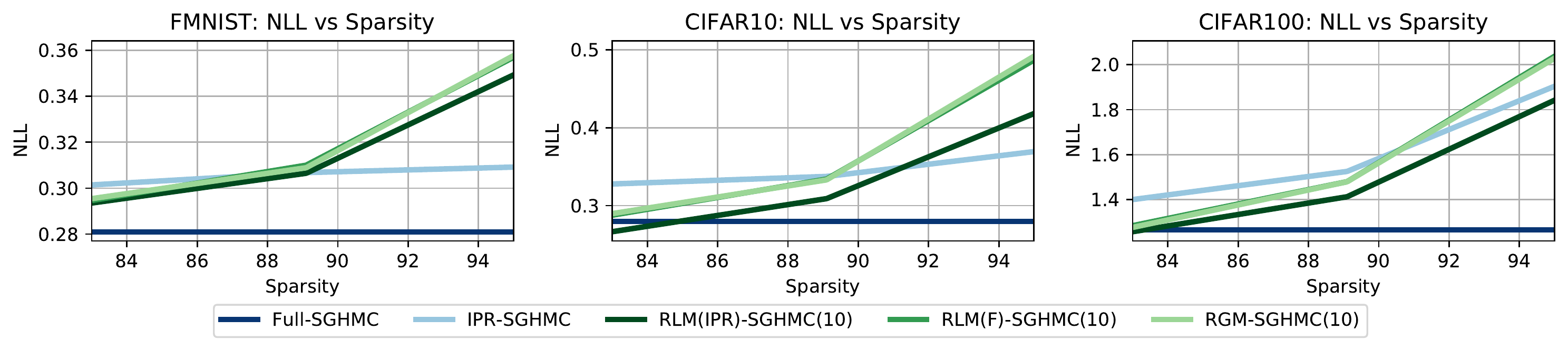}
\caption{Performance of SGHMC applied in random sparse substructures compared to SGHMC applied to full models.}
\label{fig:exp3}
\end{figure*}

\begin{wraptable}{r}{0.45\textwidth}
\small
\center
\caption{Total training time ($\downarrow$) for CIFAR10.}
\resizebox{0.45\textwidth}{!}{
\begin{tabular}{llcr}
\toprule
           Method & Chains & Sparsity &  Time (s) \\
\midrule
         IPR-SGHMC & 1 &    83\% &         19,244 \\
         IPR-SGHMC & 1 &    89\% &         22,866 \\
         IPR-SGHMC & 1 &    95\% &         30,111 \\
        RLM(F)/RGM-SGHMC & 1  &    - &          2,943 \\
        RLM(F)/RGM-SGHMC & 5  &       - &         12,452 \\
       RLM(F)/RGM-SGHMC & 10  &       - &         24,338  \\
\bottomrule
\end{tabular}
}
\label{tab:exp3b} 
\end{wraptable}

\textbf{Experiment 2: How does SGHMC within random sparse substructures compare to SGHMC within optimized sparse substructures?} While prior work on optimization-based learning indicates that directly learning high-quality models in compact or sparse models works less well than iterative pruning methods, iterative pruning methods are extremely slow to execute due to the large number of epochs needed to achieve high rates of sparsity. In this experiment we investigate the performance of different approaches to random generation of sparse substructures including random layer-wise masking with fixed sparsity rates per level (RLM(F)), as well as random global masking (RGM). We compare these against Full-SGHMC and IPR-SGHMC. We also compare to a random layer-wise masking variant that uses the per-layer sparsity levels obtained from iterative pruning with rewinding (RLM(IPR)). This approaches matches the per-layer sparsity level of the substructure identified by iterative pruning while randomizing the actual substructure. Since producing random substructures requires drastically reduced computation compared to iterative pruning, we sample several random substructures and run separate chains within each. 

In Figure \ref{fig:exp3} we show the negative log likelihood results based on randomly sampling 10 substructures and running 10 separate chains. We divide the total sample budget evenly among the 10 chains. Surprisingly, the use of random substructures and parallel chains performs as well as or better than the use of a single SGHMC chain with a sparse sub-structure using iterative pruning at lower levels of sparsity. At higher levels of sparsity, iterative pruning outperforms random substructures with parallel chains. We also see that matching the per-layer sparsity levels of iterative pruning does produce improvements over uniform per-layer sampling.

\begin{wrapfigure}{r}{0.55\textwidth}
\begin{subfigure}{.27\textwidth}
  \centering
  \includegraphics[width=1.\columnwidth]{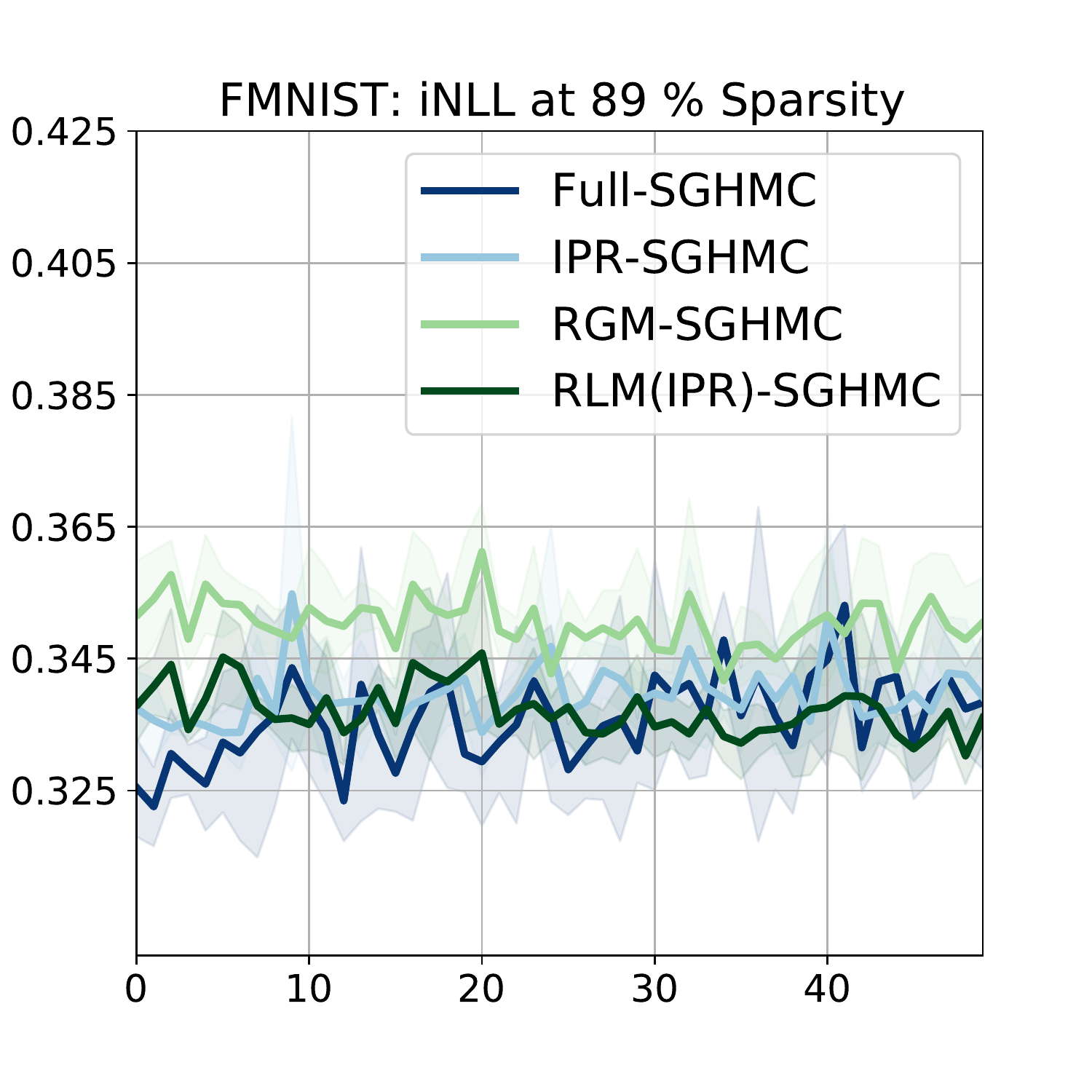}
\end{subfigure}
\begin{subfigure}{.27\textwidth}
  \centering
  \includegraphics[width=1.\columnwidth]{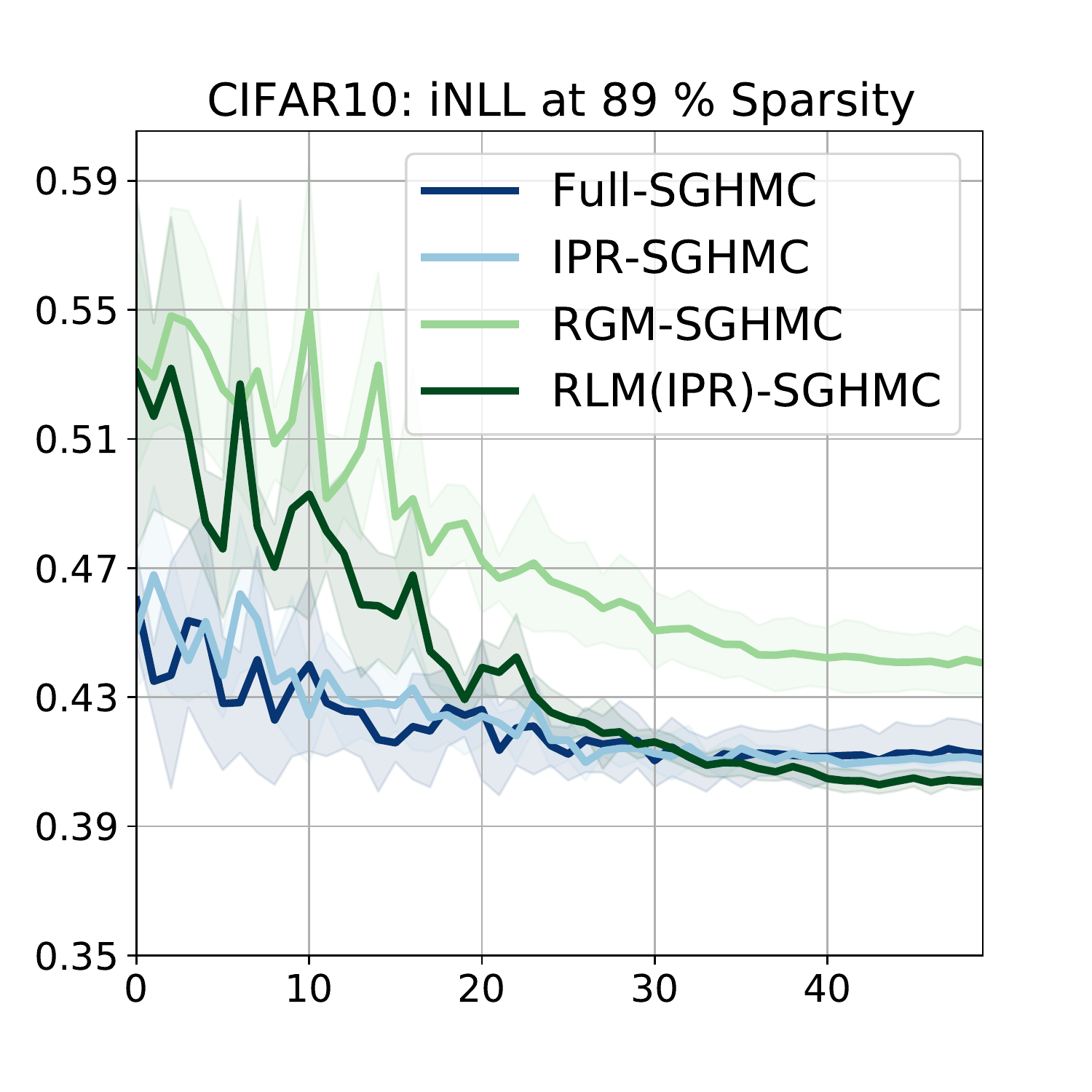}
\end{subfigure}
\caption{Plots of iNLL for FMNIST and CIFAR10 corresponding to sparsity level 89\%. We also include the full model as a baseline. 
}
\label{fig:acf_inll_89_cifar10_mnist}
\end{wrapfigure}

In Table \ref{tab:exp3b}, we show the total training time for several of the methods investigated in this experiment. As we can see, the time required by the IPR-SGHMC method is very high due to the need to run  iterative pruning to derive the sparsity mask before sampling can run. The training time also increases with the desired level of sparsity. Due to the fact that our masking implementation of SGHMC does not realize a speedup due to sparsity when sampling, the time required to run the random layer-wise masking approach with fixed sparsity per level (RLM(F)-SGHMC) and the random global masking approach (RGM-SGHMC) are the same and independent of the sparsity level (including sampling in the fully dense model). As we can see, running $10$ SGHMC chains at any level of sparsity requires total time approximately equal to running iterative pruning followed by one chain at the 89\% sparsity level. However, the 10 RLM(F)-SGHMC chains can be run completely in parallel within an actual wall-clock period equal to the total time to run one chain. Overall, these results indicate that the use of iterative pruning appears to be unnecessary at lower sparsity levels so long as multiple chains are used to offset the selection of random substructures. It also suggests a number of other interesting possibilities such as running iterative pruning to further sparsify a randomly selected sparse structure, which we leave to future work. Additional results from this experiment are presented in  Appendix \ref{app:additional_results} (Figures \ref{fig:app-exp3-1}-\ref{fig:app-exp3-3}).


\begin{wraptable}{r}{0.45\textwidth}
\small
\centering
\caption{Mean ESS ($\uparrow$) on CIFAR10.}
\label{tab:ess-ipr}
\resizebox{0.45\textwidth}{!}{
\begin{tabular}{lccc}
\toprule
Sparsity  &0 \% & 89\% & 95\% \\\midrule
IPR-SGHMC   & 39.8224 & 41.5218 & 42.0768\\
RLM(IPR)-SGHMC  & - & 38.5486 & 38.6837\\
RGM-SGHMC & - & 37.7584 & 37.9131\\
\bottomrule
\end{tabular}

}
\end{wraptable}

\textbf{Experiment 3: How does sampling within different types of sparse structures affect the underlying SGHMC Markov chain?} In this experiment, we examine the structure of the SGHMC Markov chains that result from the use of different types of sparse structures. Prior observations regarding the difficulty of optimizing compact and sparse models suggests that mixing in a gradient-based sampler such as SGHMC might degrade in sparse models. However, the size of the space that the sampler operates in is also drastically reduced at high sparsity rates, which we might expect to improve mixing. The overall effect of parameter sparsity on MCMC methods has not been previously investigated. 

Figure \ref{fig:acf_inll_89_cifar10_mnist} shows the instantaneous negative log likelihood (iNLL) on the test set for several methods applied to the FMNIST and CIFAR10 data sets. Based on the iNLL plot, the burn-in time does appear to vary by the method on CIFAR10 with random sub-substructures requiring longer to burn in than optimized substructures and the full model. This observation may be due to the use of an optimization method to select the starting state of each chain, with the iterative pruning optimization process resulting in better initial states than the application of standard optimization methods to random sparse structures. However, all chains appear to be approximately stationary by 50 sampling epochs. On FMNIST, there appears to be much less sensitivity with respect to substructures and all chains burn in quickly. To assess mixing of the MCMC chain, we compute the ESS for each of the posterior ensembles. The mean ESS values are presented in Table \ref{tab:ess-ipr}. The results presented in Table \ref{tab:ess-ipr} suggest no significant difference in ESS values as we increase the sparsity. Furthermore, as we move away from IPR based methods to random methods, we again observe no significant change in ESS values. Thus, while there is no significant improvement in ESS values from moving to dense ensembles to sparse ensembles, there's also no decrease in the ESS values. This is promising as it shows that mixing in sparse subspaces is no worse than the original space. Additionally, we also provide ACF plots for the iNLL from the 51st epoch to the 100th epoch (i.e. post burn-in) for each chain. These plots, along with the full set of iNLL plots are presented in Appendix \ref{app:additional_results}  (Figs \ref{fig:ACF_89} - \ref{fig:ACF_95_CIFAR100}).



\begin{figure*}[t]
\center
\includegraphics[width=\textwidth]{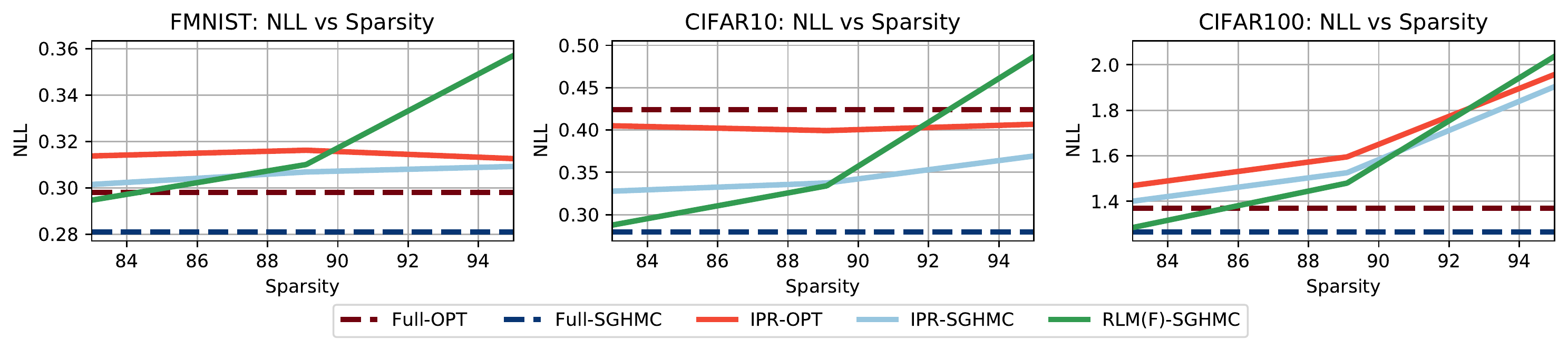}
\caption{Performance of SGHMC and optimization-based learning  on the FMNIST, CIFAR10 and CIFAR100 datasets.}
\label{fig:exp4}
\end{figure*}

\textbf{Experiment 4: How does SGHMC compare to optimization-based learning of sparse substructures?} In this experiment, we compare full models using both optimization-based learning and SGHMC to sparse models learned using both optimization and SGHMC. We also compare to randomly-selected substructures. The results are presented in Figure \ref{fig:exp4}. We again focus on iterative pruning with rewinding as the optimization-based method for deriving sparse structures. We focus on negative log likelihood as the performance measure. First, as we would expect, SGHMC out-performs optimization-based learning in the full models on all three data sets. Second, we note that SGHMC with the IPR-derived sparse structures also consistently out-performs the final optimized parameter values found for those same structures for all data sets and at all sparsity levels. 

However, the relationship between the performance of the optimized models and SGHMC in sparse substructures is more complex. As we can see, at these sparsity levels, SGHMC under-performs Full-OPT on both FMNIST and CIFAR100. Note that the random layer-wise sparsity method shown uses 10 chains and as observed previously, this leads to improvements in performance over the IPR-based SGHMC approach at lower sparsity levels. This approach does at least slightly out-perform optimization applied to the full model on all data sets. Finally, we observe that the IPR-based SGHMC method is able to out-perform the optimization-based full model on the CIFAR10 data set even at the highest level of sparsity. The full set of results is given in Appendix \ref{app:additional_results} (Fig \ref{fig:app-exp4}). The key takeaway here is that we can obtain useful trade-offs by opting with parallel chain SGHMC on sparse substructures against SGHMC/SGD on full models, leading to train and test time savings as well as storage savings. 

\section{Conclusions and Future Directions}
In this paper, we have explored the open question of how stochastic gradient MCMC methods perform when restricted to performing inference within different types of sparse sub-structures. We derive multiple important conclusions from our experimental results. First, from the standpoint of classification performance, our results show that both single chains run within single substructures derived using iterative pruning methods and multiple chains run on multiple randomly selected sparse substructures can lead to effective trade-offs between posterior storage requirements, prediction time, and predictive performance compared to a full posterior ensemble for some model/data set combinations. Perhaps the most surprising result is that at lower levels of sparsity, running multiple MCMC chains on randomly selected sub-structures performs very similarly to iterative pruning approaches that can not be parallelized, and thus require drastically higher training times. However, iterative pruning-based structures do appear to have an edge on randomly selected sparse structures in terms of predictive performance at higher sparsity rates. They also appear to exhibit better mean expected sample size and require less burn-in time. A key direction for future research is the development and evaluation of optimization methods for selecting sparse structures with more modest run times than iterative pruning strategies. Additional future directions include investigating the compression of sparse posterior ensembles using distillation methods and considering the composition of sparse substructures with subspace inference to further improve sampling efficiency.

\subsubsection*{Acknowledgments}

This work was partially supported by the US Army Research Laboratory under cooperative agreement W911NF-17-2-0196. The views and conclusions contained  in  this  document  are  those  of  the  authors  and  should  not  be interpreted as representing the official policies, either expressed  or  implied,  of  the  Army  Research  Laboratory  or  the  US  government.

\bibliographystyle{abbrvnat}
\bibliography{references}

\begin{thebibliography}{41}
\providecommand{\natexlab}[1]{#1}
\providecommand{\url}[1]{\texttt{#1}}
\expandafter\ifx\csname urlstyle\endcsname\relax
  \providecommand{\doi}[1]{doi: #1}\else
  \providecommand{\doi}{doi: \begingroup \urlstyle{rm}\Url}\fi

\bibitem[Alvarez and Salzmann(2016)]{alvarez2016learning}
J.~M. Alvarez and M.~Salzmann.
\newblock Learning the number of neurons in deep networks.
\newblock In \emph{NeurIPS}, 2016.

\bibitem[Balan et~al.(2015)Balan, Rathod, Murphy, and
  Welling]{balan2015bayesian}
A.~K. Balan, V.~Rathod, K.~P. Murphy, and M.~Welling.
\newblock Bayesian dark knowledge.
\newblock In \emph{NeurIPS}, 2015.

\bibitem[Bingham et~al.(2019)Bingham, Chen, Jankowiak, Obermeyer, Pradhan,
  Karaletsos, Singh, Szerlip, Horsfall, and Goodman]{bingham2019pyro}
E.~Bingham, J.~P. Chen, M.~Jankowiak, F.~Obermeyer, N.~Pradhan, T.~Karaletsos,
  R.~Singh, P.~A. Szerlip, P.~Horsfall, and N.~D. Goodman.
\newblock Pyro: Deep universal probabilistic programming.
\newblock \emph{J. Mach. Learn. Res.}, 20:\penalty0 28:1--28:6, 2019.
\newblock URL \url{http://jmlr.org/papers/v20/18-403.html}.

\bibitem[Blundell et~al.(2015)Blundell, Cornebise, Kavukcuoglu, and
  Wierstra]{blundell2015weight}
C.~Blundell, J.~Cornebise, K.~Kavukcuoglu, and D.~Wierstra.
\newblock Weight uncertainty in neural networks.
\newblock \emph{arXiv:1505.05424}, 2015.

\bibitem[Casella and George(1992)]{casella1992explaining}
G.~Casella and E.~I. George.
\newblock Explaining the gibbs sampler.
\newblock \emph{The American Statistician}, 46\penalty0 (3):\penalty0 167--174,
  1992.

\bibitem[Chen et~al.(2014)Chen, Fox, and Guestrin]{Chen2014StochasticGH}
T.~Chen, E.~B. Fox, and C.~Guestrin.
\newblock Stochastic gradient hamiltonian monte carlo.
\newblock In \emph{ICML}, 2014.

\bibitem[Chib and Greenberg(1995)]{chib1995understanding}
S.~Chib and E.~Greenberg.
\newblock Understanding the metropolis-hastings algorithm.
\newblock \emph{The american statistician}, 49\penalty0 (4):\penalty0 327--335,
  1995.

\bibitem[Depeweg et~al.(2017)Depeweg, Hern{\'a}ndez-Lobato, Doshi-Velez, and
  Udluft]{Depeweg2017DecompositionOU}
S.~Depeweg, J.~M. Hern{\'a}ndez-Lobato, F.~Doshi-Velez, and S.~Udluft.
\newblock Decomposition of uncertainty for active learning and reliable
  reinforcement learning in stochastic systems.
\newblock \emph{ArXiv}, abs/1710.07283, 2017.

\bibitem[Duane et~al.(1987)Duane, Kennedy, Pendleton, and
  Roweth]{duane1987hybrid}
S.~Duane, A.~D. Kennedy, B.~J. Pendleton, and D.~Roweth.
\newblock Hybrid monte carlo.
\newblock \emph{Physics letters B}, 195\penalty0 (2):\penalty0 216--222, 1987.

\bibitem[Frankle and Carbin(2019)]{Frankle2018TheLT}
J.~Frankle and M.~Carbin.
\newblock The lottery ticket hypothesis: Training pruned neural networks.
\newblock \emph{ICLR}, 2019.

\bibitem[Gal and Ghahramani(2016)]{gal2016dropout}
Y.~Gal and Z.~Ghahramani.
\newblock Dropout as a bayesian approximation: Representing model uncertainty
  in deep learning.
\newblock In \emph{ICML}, 2016.

\bibitem[Galassi et~al.(2003)Galassi, Davies, Theiler, Gough, Jungman, Alken,
  Booth, Rossi, and Ulerich]{galassi2003gnu}
M.~Galassi, J.~Davies, J.~Theiler, B.~Gough, G.~Jungman, P.~Alken, M.~Booth,
  F.~Rossi, and R.~Ulerich.
\newblock Gnu scientific library.
\newblock \emph{Reference Manual. Edition 1.4, for GSL Version 1.4}, 2003.

\bibitem[Girolami and Calderhead(2011)]{girolami2011riemann}
M.~Girolami and B.~Calderhead.
\newblock Riemann manifold langevin and hamiltonian monte carlo methods.
\newblock \emph{Journal of the Royal Statistical Society: Series B (Statistical
  Methodology)}, 73\penalty0 (2):\penalty0 123--214, 2011.

\bibitem[Goodfellow et~al.(2015)Goodfellow, Shlens, and
  Szegedy]{Goodfellow2015ICLR}
I.~J. Goodfellow, J.~Shlens, and C.~Szegedy.
\newblock Explaining and harnessing adversarial examples.
\newblock In \emph{ICLR}, 2015.

\bibitem[Graves(2011)]{graves2011practical}
A.~Graves.
\newblock Practical variational inference for neural networks.
\newblock In \emph{NeurIPS}, 2011.

\bibitem[Guo et~al.(2017)Guo, Pleiss, Sun, and Weinberger]{guo2017calibration}
C.~Guo, G.~Pleiss, Y.~Sun, and K.~Q. Weinberger.
\newblock On calibration of modern neural networks.
\newblock In \emph{ICML}, pages 1321--1330, 2017.

\bibitem[Han et~al.(2015)Han, Pool, Tran, and Dally]{Han2015LearningBW}
S.~Han, J.~Pool, J.~Tran, and W.~J. Dally.
\newblock Learning both weights and connections for efficient neural networks.
\newblock In \emph{NeurIPS}, 2015.

\bibitem[Hassibi et~al.(1993)Hassibi, Stork, and Wolff]{Hassibi1993OptimalBS}
B.~Hassibi, D.~G. Stork, and G.~J. Wolff.
\newblock Optimal brain surgeon and general network pruning.
\newblock \emph{IEEE International Conference on Neural Networks}, 1993.

\bibitem[He et~al.(2017)He, Zhang, and Sun]{He2017ChannelPF}
Y.~He, X.~Zhang, and J.~Sun.
\newblock Channel pruning for accelerating very deep neural networks.
\newblock \emph{ICCV}, 2017.

\bibitem[Izmailov et~al.(2019)Izmailov, Maddox, Kirichenko, Garipov, Vetrov,
  and Wilson]{Izmailov2019SubspaceIF}
P.~Izmailov, W.~J. Maddox, P.~Kirichenko, T.~Garipov, D.~P. Vetrov, and A.~G.
  Wilson.
\newblock Subspace inference for bayesian deep learning.
\newblock In \emph{UAI}, 2019.

\bibitem[Jaakkola and Jordan(2000)]{jaakkola2000bayesian}
T.~S. Jaakkola and M.~I. Jordan.
\newblock Bayesian parameter estimation via variational methods.
\newblock \emph{Statistics and Computing}, 10\penalty0 (1):\penalty0 25--37,
  2000.

\bibitem[Jordan et~al.(1999)Jordan, Ghahramani, Jaakkola, and
  Saul]{jordan1999introduction}
M.~I. Jordan, Z.~Ghahramani, T.~S. Jaakkola, and L.~K. Saul.
\newblock An introduction to variational methods for graphical models.
\newblock \emph{Machine learning}, 37\penalty0 (2):\penalty0 183--233, 1999.

\bibitem[Kernighan and Ritchie(2006)]{kernighan2006c}
B.~W. Kernighan and D.~M. Ritchie.
\newblock \emph{The C programming language}.
\newblock 2006.

\bibitem[Krizhevsky et~al.(2009)Krizhevsky, Hinton,
  et~al.]{krizhevsky2009learning}
A.~Krizhevsky, G.~Hinton, et~al.
\newblock Learning multiple layers of features from tiny images.
\newblock Technical report, 2009.

\bibitem[Krizhevsky et~al.(2012)Krizhevsky, Sutskever, and
  Hinton]{krizhevsky2012imagenet}
A.~Krizhevsky, I.~Sutskever, and G.~E. Hinton.
\newblock Imagenet classification with deep convolutional neural networks.
\newblock In \emph{NeurIPS}, 2012.

\bibitem[LeCun et~al.(1989)LeCun, Denker, and Solla]{LeCun1989OptimalBD}
Y.~LeCun, J.~S. Denker, and S.~A. Solla.
\newblock Optimal brain damage.
\newblock In \emph{NeurIPS}, 1989.

\bibitem[Li et~al.(2017)Li, Park, and Tang]{li2017enabling}
S.~Li, J.~Park, and P.~T.~P. Tang.
\newblock Enabling sparse winograd convolution by native pruning, 2017.

\bibitem[Louizos et~al.(2017)Louizos, Ullrich, and
  Welling]{Louizos2017BayesianCF}
C.~Louizos, K.~Ullrich, and M.~Welling.
\newblock Bayesian compression for deep learning.
\newblock \emph{ArXiv}, abs/1705.08665, 2017.

\bibitem[Neal(1996)]{Neal:1996:BLN:525544}
R.~M. Neal.
\newblock \emph{Bayesian Learning for Neural Networks}.
\newblock Springer-Verlag, 1996.

\bibitem[Neal(2003)]{neal2003slice}
R.~M. Neal.
\newblock Slice sampling.
\newblock \emph{Annals of statistics}, pages 705--741, 2003.

\bibitem[Park et~al.(2016)Park, Li, Wen, Li, Chen, and
  Dubey]{Park2016HolisticSF}
J.~Park, S.~Li, W.~Wen, H.~H. Li, Y.~Chen, and P.~Dubey.
\newblock Holistic sparsecnn: Forging the trident of accuracy, speed, and size.
\newblock \emph{ArXiv}, abs/1608.01409, 2016.

\bibitem[Park et~al.(2017)Park, Li, Wen, Tang, Li, Chen, and
  Dubey]{park2016faster}
J.~Park, S.~Li, W.~Wen, P.~T.~P. Tang, H.~Li, Y.~Chen, and P.~Dubey.
\newblock Faster cnns with direct sparse convolutions and guided pruning.
\newblock In \emph{ICLR}, 2017.

\bibitem[Robert and Casella(2013)]{robert2013monte}
C.~Robert and G.~Casella.
\newblock \emph{Monte Carlo statistical methods}.
\newblock Springer Science \& Business Media, 2013.

\bibitem[Smith and Roberts(1993)]{smith1993bayesian}
A.~F. Smith and G.~O. Roberts.
\newblock Bayesian computation via the gibbs sampler and related markov chain
  monte carlo methods.
\newblock \emph{Journal of the Royal Statistical Society: Series B
  (Methodological)}, 55\penalty0 (1):\penalty0 3--23, 1993.

\bibitem[Vadera et~al.(2020{\natexlab{a}})Vadera, Cobb, Jalaian, and
  Marlin]{Vadera2020URSABench}
M.~P. Vadera, A.~D. Cobb, B.~Jalaian, and B.~M. Marlin.
\newblock Ursabench: Comprehensive benchmarking of approximate bayesian
  inference methods for deep neural networks.
\newblock \emph{arXiv preprint arXiv:2007.04466}, 2020{\natexlab{a}}.
\newblock URL \url{https://github.com/reml-lab/URSABench}.

\bibitem[Vadera et~al.(2020{\natexlab{b}})Vadera, Jalaian, and
  Marlin]{Vadera2020GeneralizedBP}
M.~P. Vadera, B.~Jalaian, and B.~M. Marlin.
\newblock Generalized bayesian posterior expectation distillation for deep
  neural networks.
\newblock In \emph{UAI}, 2020{\natexlab{b}}.

\bibitem[Welling and Teh(2011)]{welling2011bayesian}
M.~Welling and Y.~W. Teh.
\newblock Bayesian learning via stochastic gradient langevin dynamics.
\newblock In \emph{ICML}, pages 681--688, 2011.

\bibitem[Wen et~al.(2016)Wen, Wu, Wang, Chen, and Li]{wen2016learning}
W.~Wen, C.~Wu, Y.~Wang, Y.~Chen, and H.~Li.
\newblock Learning structured sparsity in deep neural networks.
\newblock In \emph{NeurIPS}, 2016.

\bibitem[Xiao et~al.(2017)Xiao, Rasul, and Vollgraf]{Xiao2017FashionMNISTAN}
H.~Xiao, K.~Rasul, and R.~Vollgraf.
\newblock Fashion-mnist: a novel image dataset for benchmarking machine
  learning algorithms.
\newblock \emph{ArXiv}, abs/1708.07747, 2017.

\bibitem[Zhang et~al.(2020)Zhang, Li, Zhang, Chen, and
  Wilson]{Zhang2020Cyclical}
R.~Zhang, C.~Li, J.~Zhang, C.~Chen, and A.~G. Wilson.
\newblock Cyclical stochastic gradient mcmc for bayesian deep learning.
\newblock In \emph{ICLR}, 2020.

\bibitem[Zhang and Ou(2018)]{Zhang2018LEARNINGSS}
Y.~Zhang and Z.~Ou.
\newblock Learning sparse structured ensembles with stochastic gradient mcmc
  sampling and network pruning.
\newblock \emph{IEEE MLSP}, 2018.

\end{thebibliography}

\clearpage
\appendix
\section{Appendix}
\subsection{SGHMC in sparse neural networks}
\label{app:sparse_sghmc}
In Algorithm \ref{sghmc_alg}, we present the SGHMC algorithm we use for neural networks with sparse sub-structure.

\begin{algorithm}[htbp]
\caption{Sparse Sub-Structure SGHMC}
\label{sghmc_alg}
\begin{algorithmic}
\State \textbf{Inputs:} $\theta^{(0)}$, $\mbf{m}$, $\mathcal{D}_{tr}$, $B$, $\alpha_{0:S}$, $\eta$, $S$ 
\State $\mbf{r} \leftarrow 0$
\For{$(0\leq s \leq S)$}
\State Select mini-batch $\mathcal{D}'_{tr}\subset \mathcal{D}_{tr}$ of size $B$
\State $\nabla \hat{U} \leftarrow -\nabla p(\theta^{(s)}|\gamma) -\frac{N}{B}\sum_{(\mbf{x},y)\in\mathcal{D}'} \nabla p(y|\mbf{x},\theta^{(s)})$
\State $\mbf{r} \leftarrow (1-\eta)\mbf{r} -\alpha_k \nabla \hat{U} + \epsilon\cdot \sqrt{2\eta\alpha_k}$,\;\;\;  $\epsilon \sim \mathcal{N}(0,I)$
\State $\theta^{(s+1)} \leftarrow \mbf{m} \odot(\theta^{(s)} + \mbf{r})$
\EndFor
\State \textbf{Return:} $[\theta^{(1)},...,\theta^{(S)}]$
\end{algorithmic}
\end{algorithm}

\subsection{Implementation Details}
\label{app:implementation_details}

In this section we include the implementation details of the experimental protocols outlined in Section \ref{sec:experiments:protocols}. For hyperparameter optimization, we started with the hyperparameters from \cite{Frankle2018TheLT} and further tuned the learning rate and scheduler to improve on their performance. We found our hyperparameters to work better.

\begin{itemize}
    \item \textbf{MLP200/FMNIST:} 
    \begin{itemize}
        \item \textbf{SGD:} For all use-cases, whether for pre-training or for IP/IPR, we use standard SGD with a learning rate of $0.01$, a weight decay of $1\times10^{-3}$, a momentum of $0.9$, and $60$ epochs. For the learning rate scheduler, we use a multi-step scheduler which reduces the learning rate by a factor of $0.1$ at epochs $20$ and $40$.
        \item \textbf{SGHMC:} For all implementations of SGHMC, we initialize the weights using SGD (as described above). We allow for a burn in of $50$ epochs and then collect the number of pre-specified samples (e.g. for one chain we collect 50 samples). The learning rate is set to a constant value of $\alpha_s = 0.01$, the friction term $\eta = 0.1$ (i.e. momentum is $0.9$), the prior precision is set to $60$ ($1\times10^{-3} \times |\mathcal{D}_{tr}|$) 
    \end{itemize}
    \item \textbf{ResNet20/CIFAR10:}
    \begin{itemize}
        \item \textbf{SGD:} We use SGD with a learning rate of $0.01$, a weight decay of $1\times10^{-4}$, a momentum of $0.9$, and $160$ epochs. For the learning rate scheduler, we use a multi-step scheduler which reduces the learning rate by a factor of $0.1$ at epochs $80$ and $120$.
        \item \textbf{SGHMC:} As for the MLP200, we initialize the weights using SGD (as described above). We allow for a burn in of $50$ epochs and then collect the number of pre-specified samples. The learning rate follows the cosine annealing scheduler with an initial learning rate of $0.2$ and a final value of $0.0$. The friction term $\eta = 0.5$ (i.e. momentum is $0.5$), the prior precision is set to $60$ ($1\times10^{-4} \times |\mathcal{D}_{tr}|$)
    \end{itemize}
    \item \textbf{ResNet20/CIFAR100:}
    \begin{itemize}
        \item \textbf{SGD:} We use the same settings as for ResNet20/CIFAR10.
        \item \textbf{SGHMC:} We use the same settings as for ResNet20/CIFAR10.
    \end{itemize}
\end{itemize}

For our paper, we use the SGHMC implementation provided by \citet{Vadera2020URSABench}. We also utilize parts of the PyTorch implementation provided by \citet{Frankle2018TheLT} for our experiments. Since the code for all our experiments is designed for a Slurm based GPU computing cluster, and can require several days to reproduce the results, we do not provide the full implementation with the supplementary material. However, the code for measure inference time at various sparsity levels for the MLP200/FMNIST setting can be found in the supplemental material. Please look at the \texttt{README.txt} file for further instructions on how to run the code in the supplemental material. 

\subsection{Evaluation Metrics} 
\label{app:methods:metrics}

In this section we describe the metrics that we apply to evaluate the performance of sampling within different sparse sub-structures. Our primary interest is to evaluate the impact of different sparsity levels and methods for selecting sparse sub-structures on the resulting posterior ensembles as well as on the SGHMC algorithm itself. In this work, our focus will be on evaluating neural network classification models. The primary metrics that we will consider are accuracy, negative log likelihood, expected calibration error, autocorrelation of the SGHMC chains and the effective sample size of the posterior ensemble. 

\noindent\textbf{{Accuracy (Acc):}} Recall that $\hat{p}(y|\mathbf{x},\mathcal{D}_{tr})$ denotes the Monte Carlo approximation to the posterior predictive distribution. We define $\hat{f}(\mathbf{x},\mathcal{D}_{tr}) = \argmax_{y'\in\mathcal{Y}} \hat{p}(y|\mathbf{x},\mathcal{D}_{tr})$ to be the corresponding classification function. We use a test set $\mathcal{D}_{te}$ to compute the accuracy of this classification function: $\mbox{Acc} = \sum_{(\mbf{x},y)\in\mathcal{D}_{te}} \left[y = \hat{f}(\mathbf{x},\mathcal{D}_{tr}) \right]$.

\noindent\textbf{{Negative Log-Likelihood (NLL):}} We compute the average negative predictive log likelihood given the posterior ensemble as: 
$\mbox{NLL} = -(1/N_{te})\sum_{(\mbf{x},y)\in\mathcal{D}_{te}} \log \hat{p}(y|\mathbf{x},\mathcal{D}_{tr})$. This measure provides more sensitivity to the quality of posterior predictive probabilities than accuracy.  

\noindent\textbf{{Expected Calibration Error (ECE):}} We compute expected calibration error of the approximate posterior predictive distribution $\hat{p}(y|\mathbf{x},\mathcal{D}_{tr})$ on the test set following the approach of \cite{guo2017calibration}. In a perfectly calibrated model, among all of the data cases predicted to be positive with probability $p$, we would expect the fraction that are true positives to be exactly $p$. The expected calibration error estimates the degree of deviation of a model's predictive probabilities from perfect calibration.

\noindent\textbf{{Autocorrelation Function (ACF):}} Autocorrelation functions are used to asses how correlated a univariate sequence with itself. The ACF is a useful tool for analyzing the correlation between functions of samples collected via a Markov chain. We select the instantaneous negative log likelihood function on the test set $\mbox{iNLL}^{(s)} = -(1/N_{te})\sum_{(\mbf{x},y)\in\mathcal{D}_{te}} \log p(y|\mathbf{x},\theta^{(s)})$ as the summary statistic as it is a convenient way to summarize high dimensional sampled parameters. If nearby samples are very similar at a given lag, then the instantaneous log likelihood values will also be similar and the ACF will be high. Samples that are approximately IID would have ACF near zero at all lags. 

\noindent\textbf{{Effective Sample Size (ESS):}} The effective sample size is used to assess the effective number of independently drawn samples materialized from a collection of Markov chains \citet[Page~499]{robert2013monte}. The ESS takes into account both the variance within and between chains and therefore summarizes the correlation between samples. High values indicate a more efficient Markov chain, where the samples are closer to being independent. As an example, completely independent samples would correspond to an ESS equal to the total number of samples collected. We use the implementation provided in Pyro to compute the ESS  \citep{bingham2019pyro}. 

\subsection{Additional Results}
\label{app:additional_results}
\begin{figure*}[t]
\center
\includegraphics[width=\textwidth]{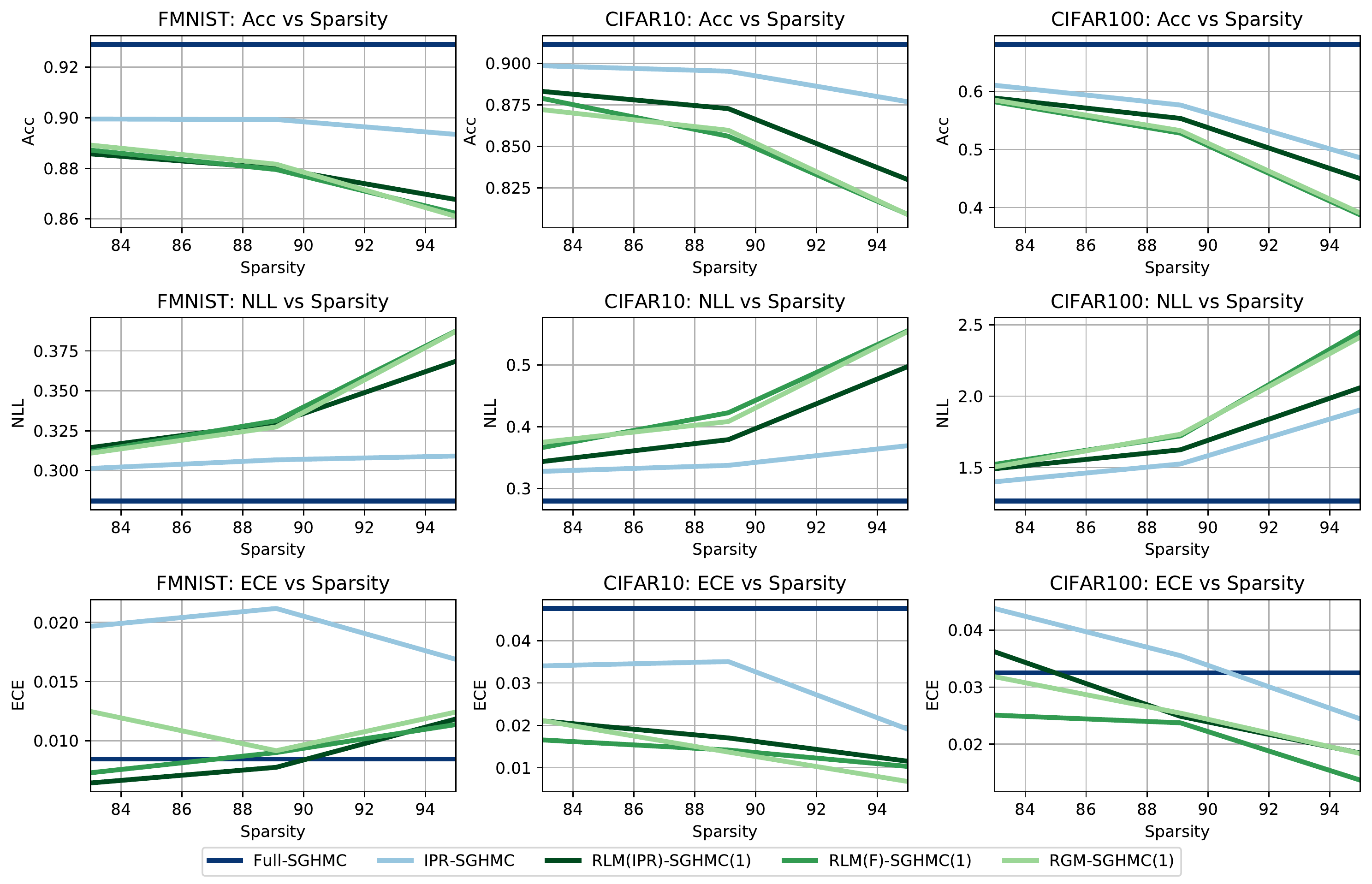}
\caption{Performance comparison of SGHMC applied in random sparse sub-structure compared to SGHMC applied to full models on the FMNIST, CIFAR10 and CIFAR100 datasets. In this figure, we sample 1 random sparse sub-structure using each for the random sparse sub-structure selection methods denoted in the legend, and apply SGHMC to it. We collect total 50 samples from the chain.}
\label{fig:app-exp3-1}
\end{figure*}

\begin{figure*}[t]
\center
\includegraphics[width=\textwidth]{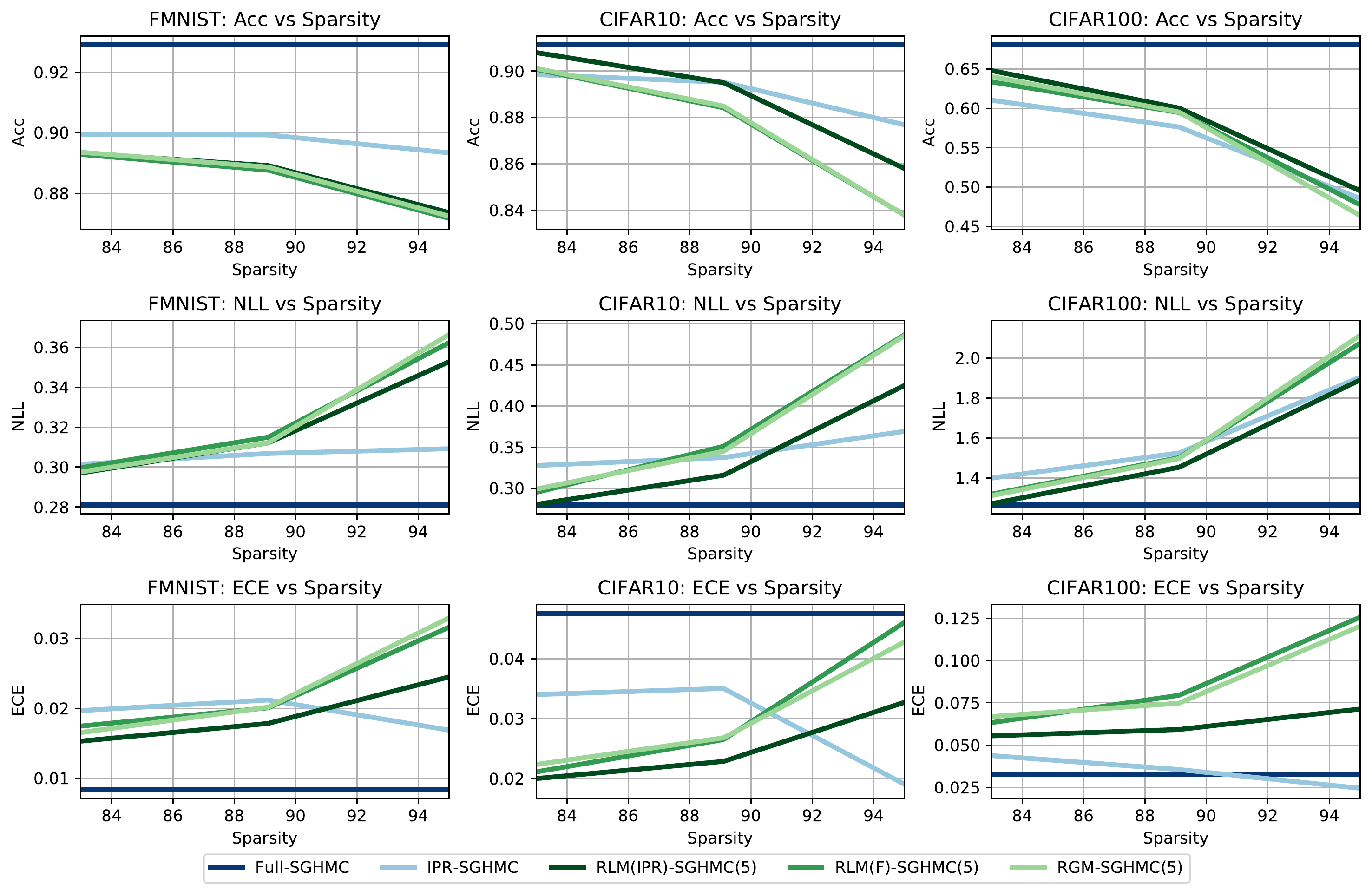}
\caption{Performance comparison of SGHMC applied in random sparse sub-structure compared to SGHMC applied to full models on the FMNIST, CIFAR10 and CIFAR100 datasets. In this figure, we sample 5 random sparse sub-structures each using the random sparse sub-structure selection methods denoted in the legend, and run 5 parallel SGHMC chains. We collect 10 samples from each chain to obtain a total of 50 samples across 5 chains.}
\label{fig:app-exp3-2}
\end{figure*}

\begin{figure*}[t]
\center
\includegraphics[width=\textwidth]{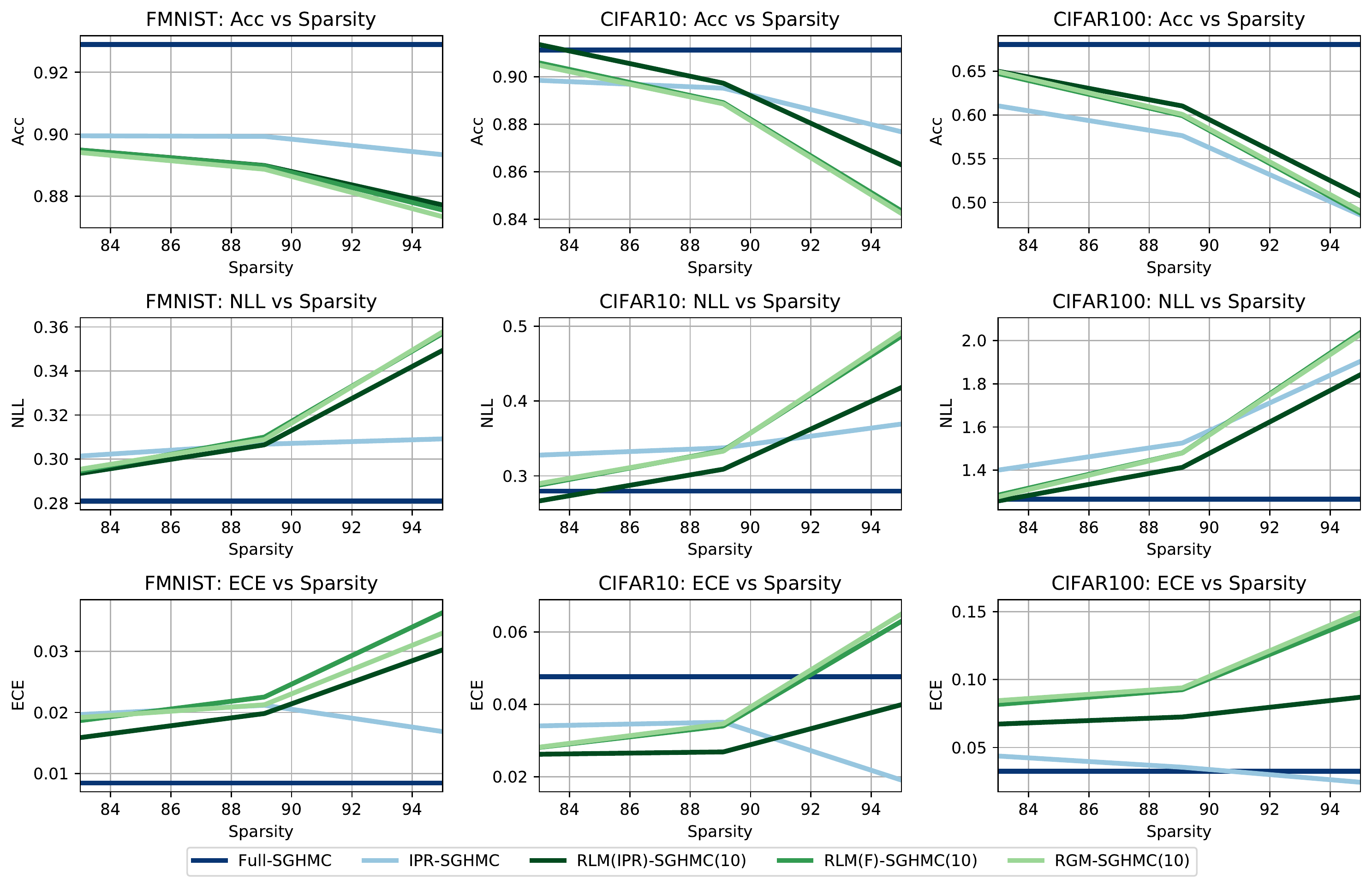}
\caption{Performance comparison of SGHMC applied in random sparse sub-structure compared to SGHMC applied to full models on the FMNIST, CIFAR10 and CIFAR100 datasets. In this figure, we sample 10 random sparse sub-structures each using the random sparse sub-structure selection methods denoted in the legend, and run 10 parallel SGHMC chains. We collect 5 samples from each chain to obtain a total of 50 samples across 10 chains.}
\label{fig:app-exp3-3}
\end{figure*}

\begin{figure}[t]
\centering
\begin{subfigure}{.49\textwidth}
  \centering
  \includegraphics[width=1.\textwidth]{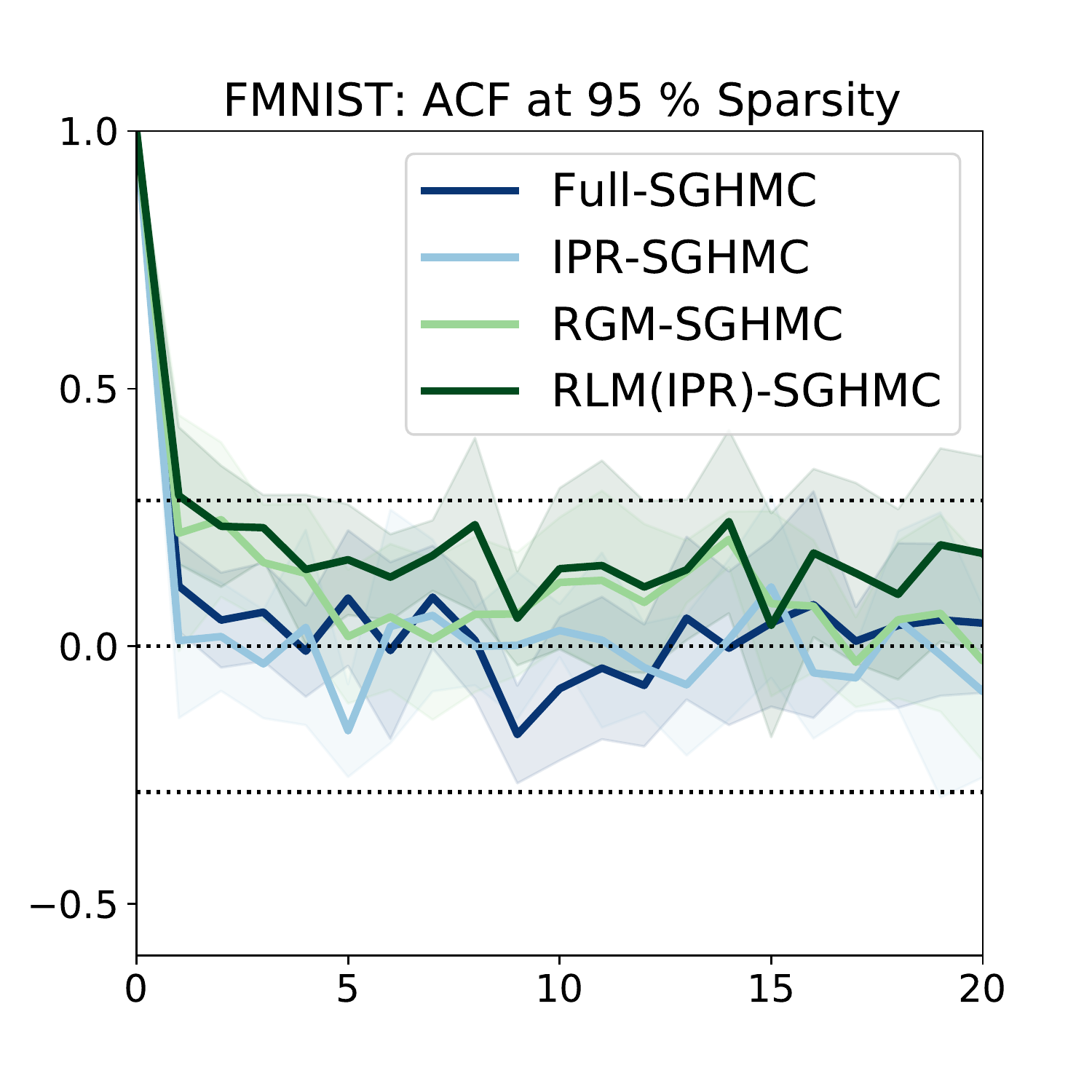}
\end{subfigure}
\begin{subfigure}{.49\textwidth}
  \centering
  \includegraphics[width=1.\textwidth]{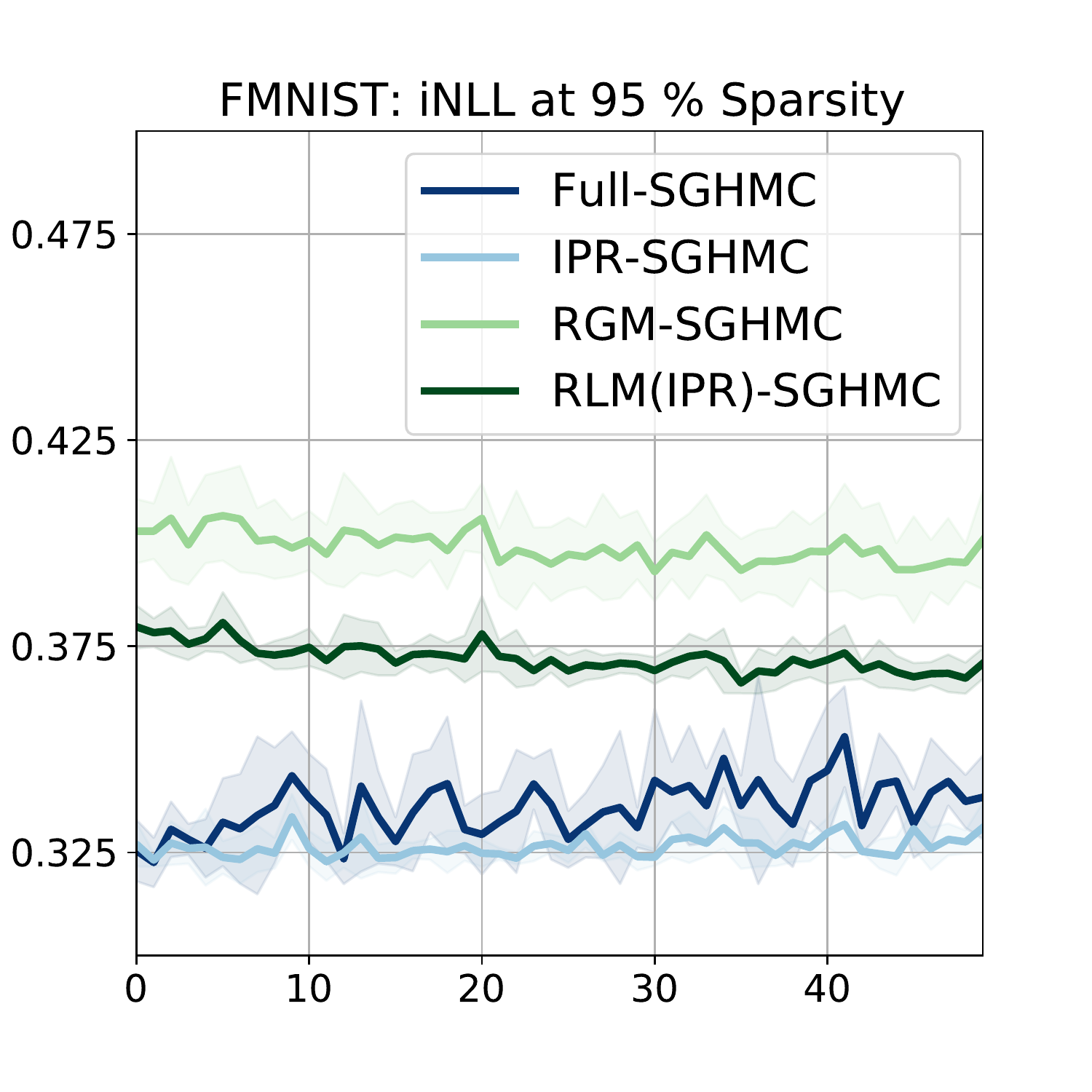}
\end{subfigure}
\begin{subfigure}{.49\textwidth}
  \centering
  \includegraphics[width=1.\textwidth]{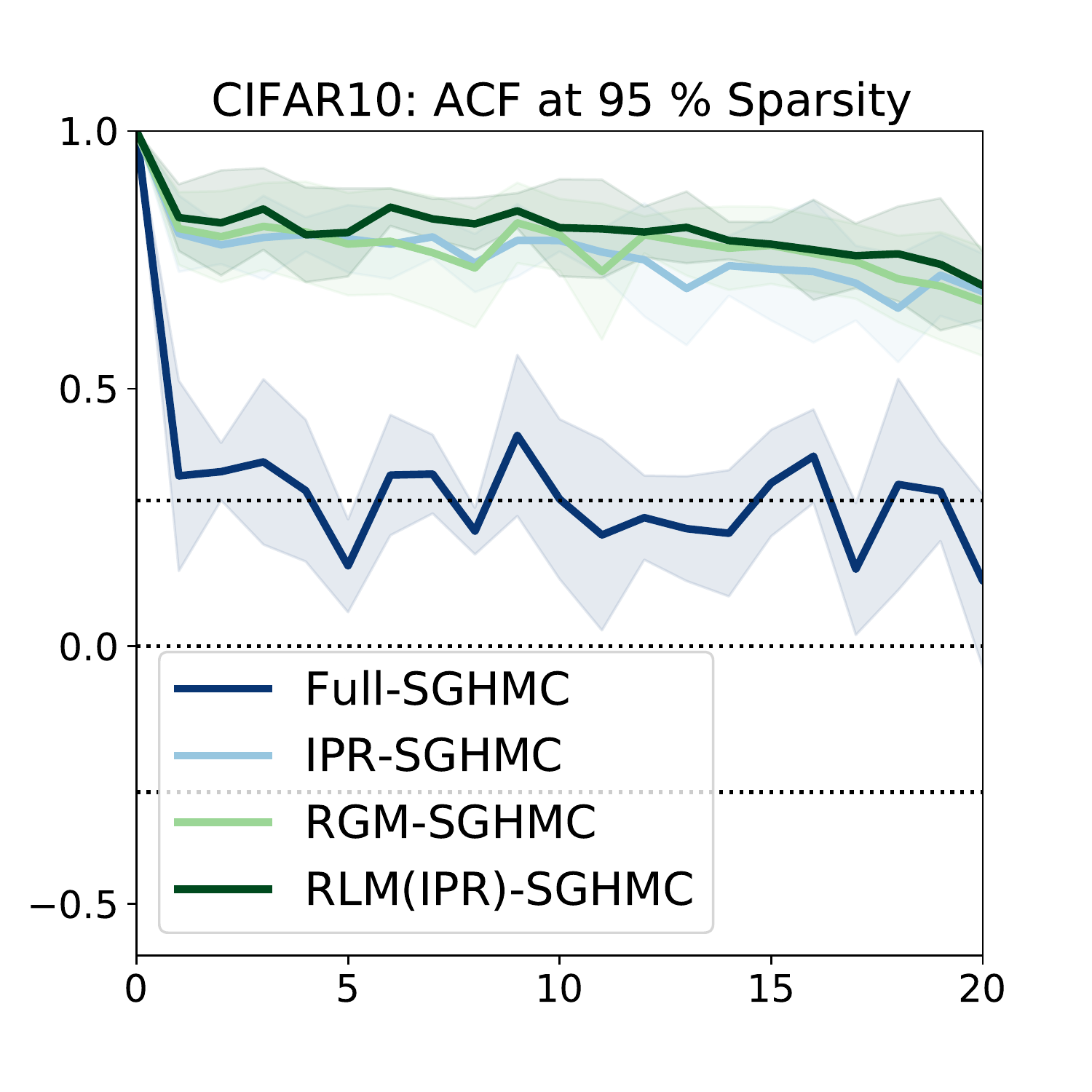}
\end{subfigure}
\begin{subfigure}{.49\textwidth}
  \centering
  \includegraphics[width=1.\textwidth]{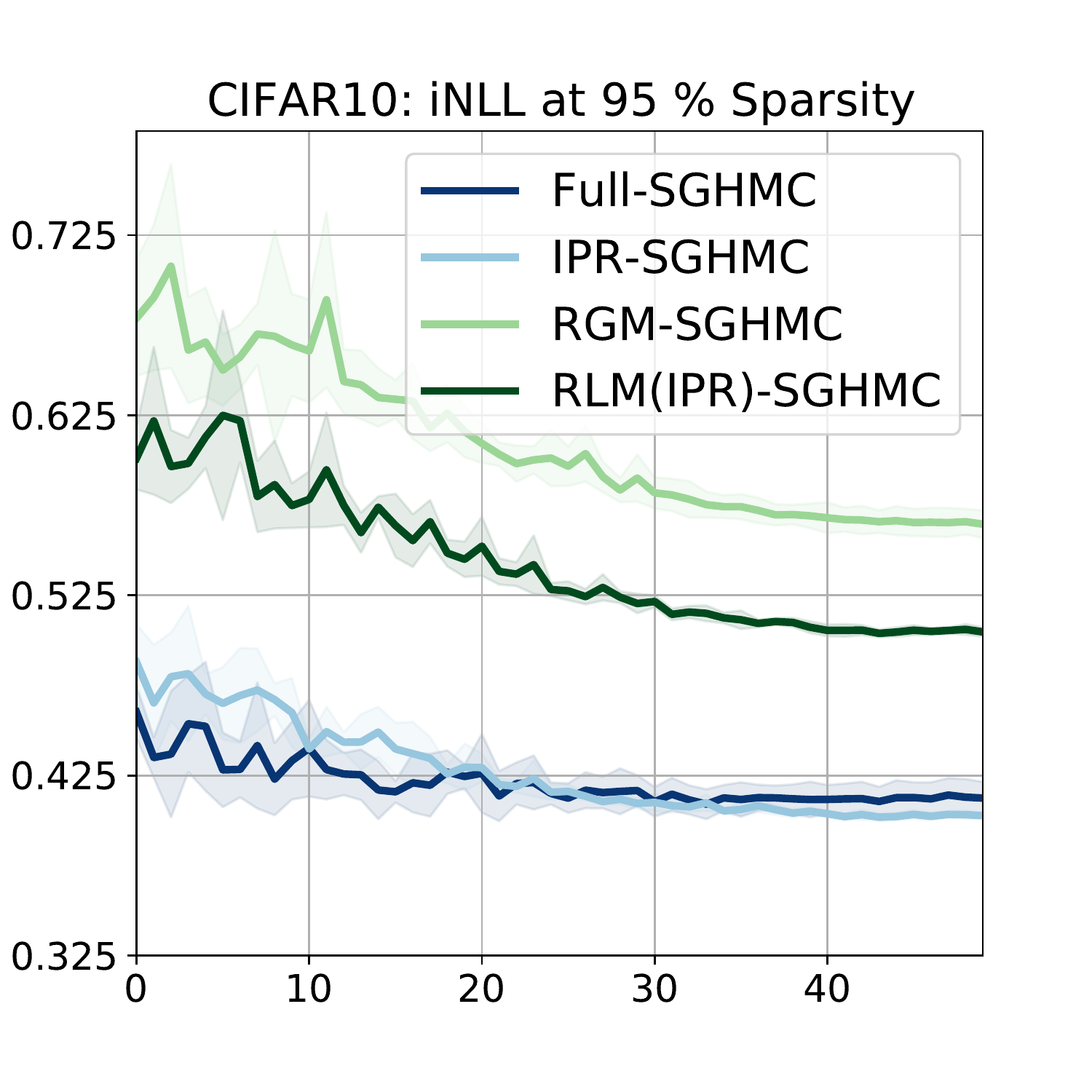}
\end{subfigure}
\caption{The top two figures show ACF and iNLL for FMNIST while the bottom two figures show ACF and iNLL for CIFAR10. The plots correspond to the case when sparsity level is set to 95\%. We also include the full model as a baseline.}
\label{fig:ACF_95}
\end{figure}

\begin{figure}[t]
\centering
\begin{subfigure}{.49\textwidth}
  \centering
  \includegraphics[width=1.\textwidth]{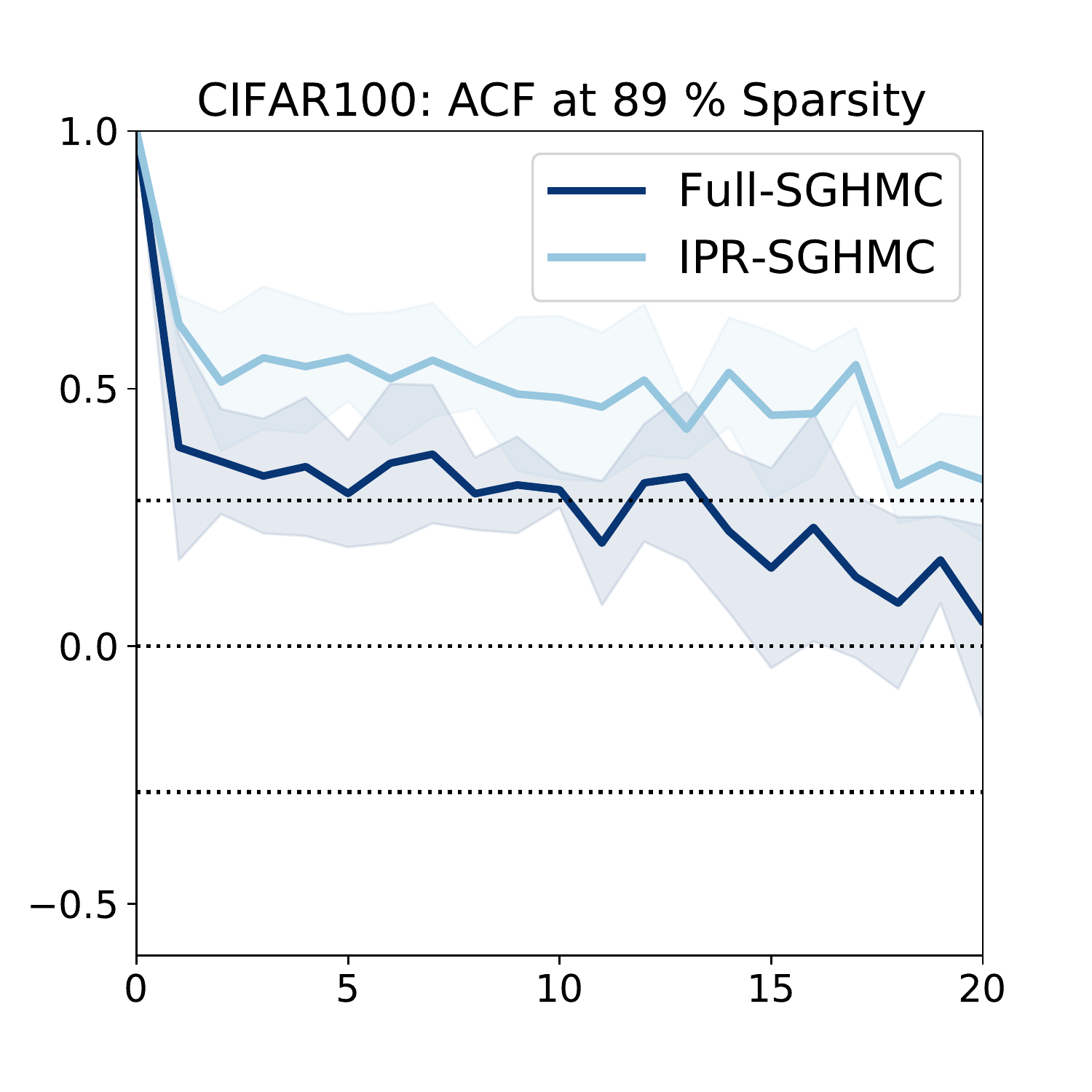}
\end{subfigure}
\begin{subfigure}{.49\textwidth}
  \centering
  \includegraphics[width=1.\textwidth]{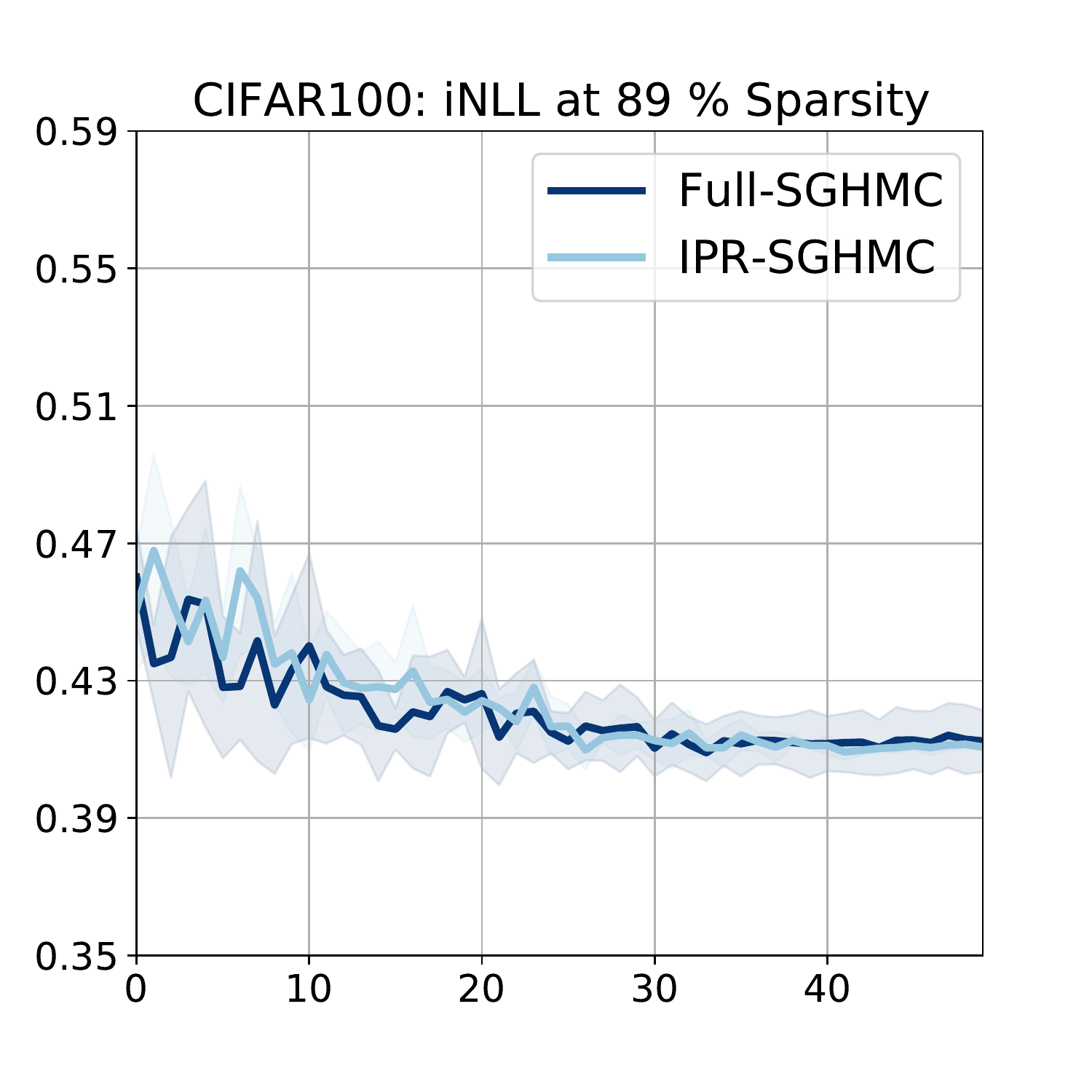}
\end{subfigure}
\begin{subfigure}{.49\textwidth}
  \centering
  \includegraphics[width=1.\textwidth]{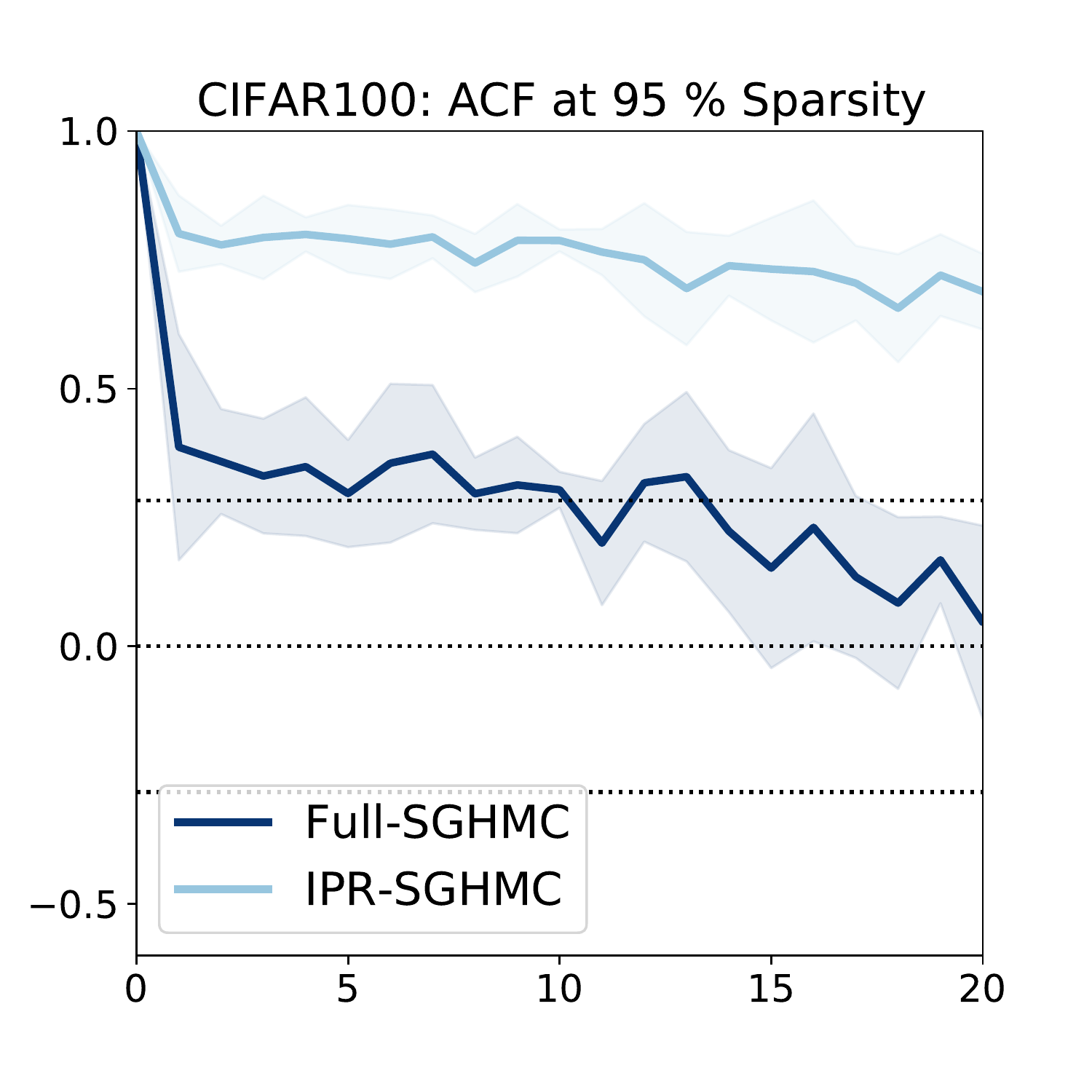}
\end{subfigure}
\begin{subfigure}{.49\textwidth}
  \centering
  \includegraphics[width=1.\textwidth]{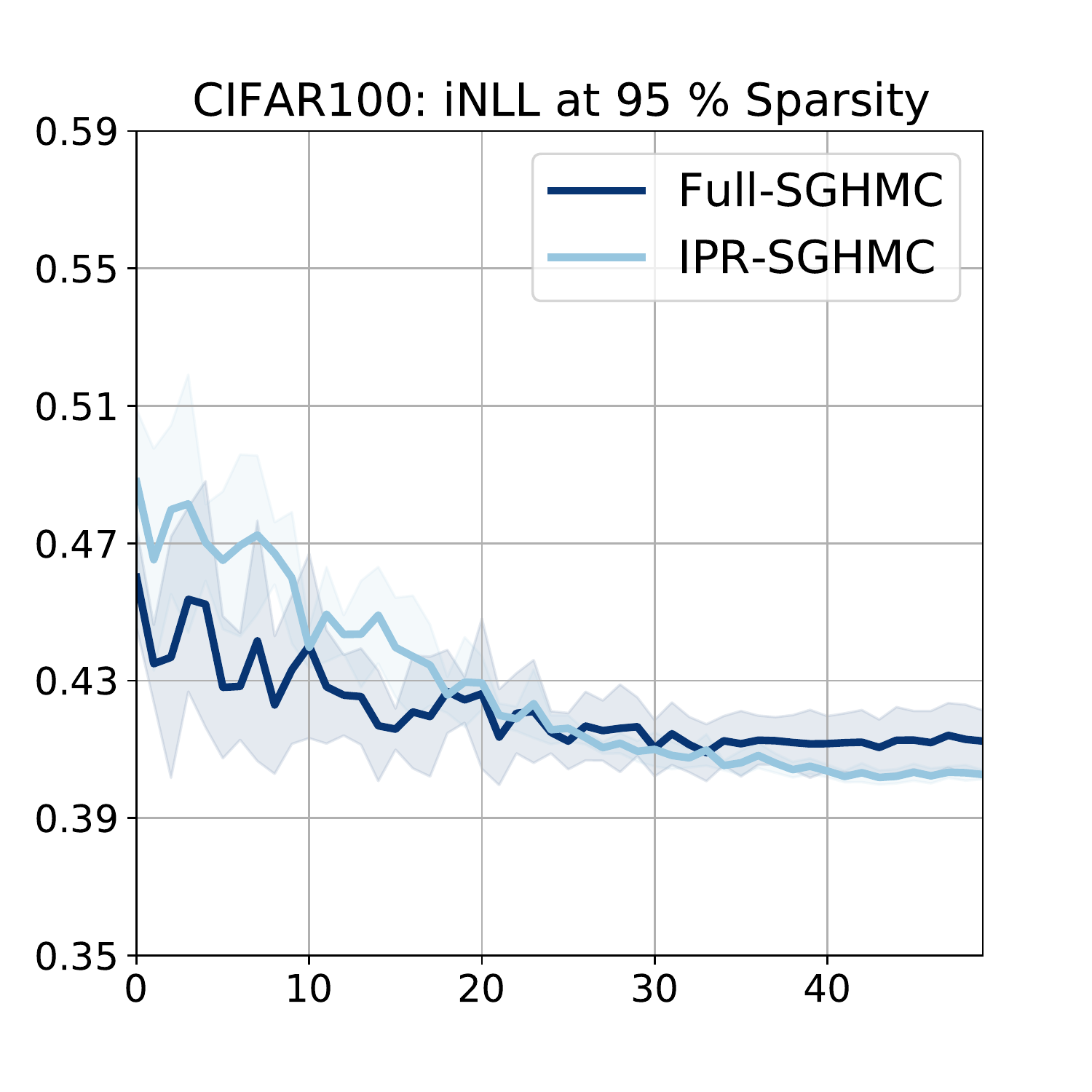}
\end{subfigure}
\caption{The top two figures show ACF and iNLL for CIFAR100 for a sparsity level of $89 \%$ while the bottom two figures show ACF and iNLL for CIFAR100 for a sparsity level of $95 \%$. We also include a baseline with sparsity rate of 0\%}
\label{fig:ACF_95_CIFAR100}
\end{figure}

\begin{figure*}[t]
\center
\includegraphics[width=\textwidth]{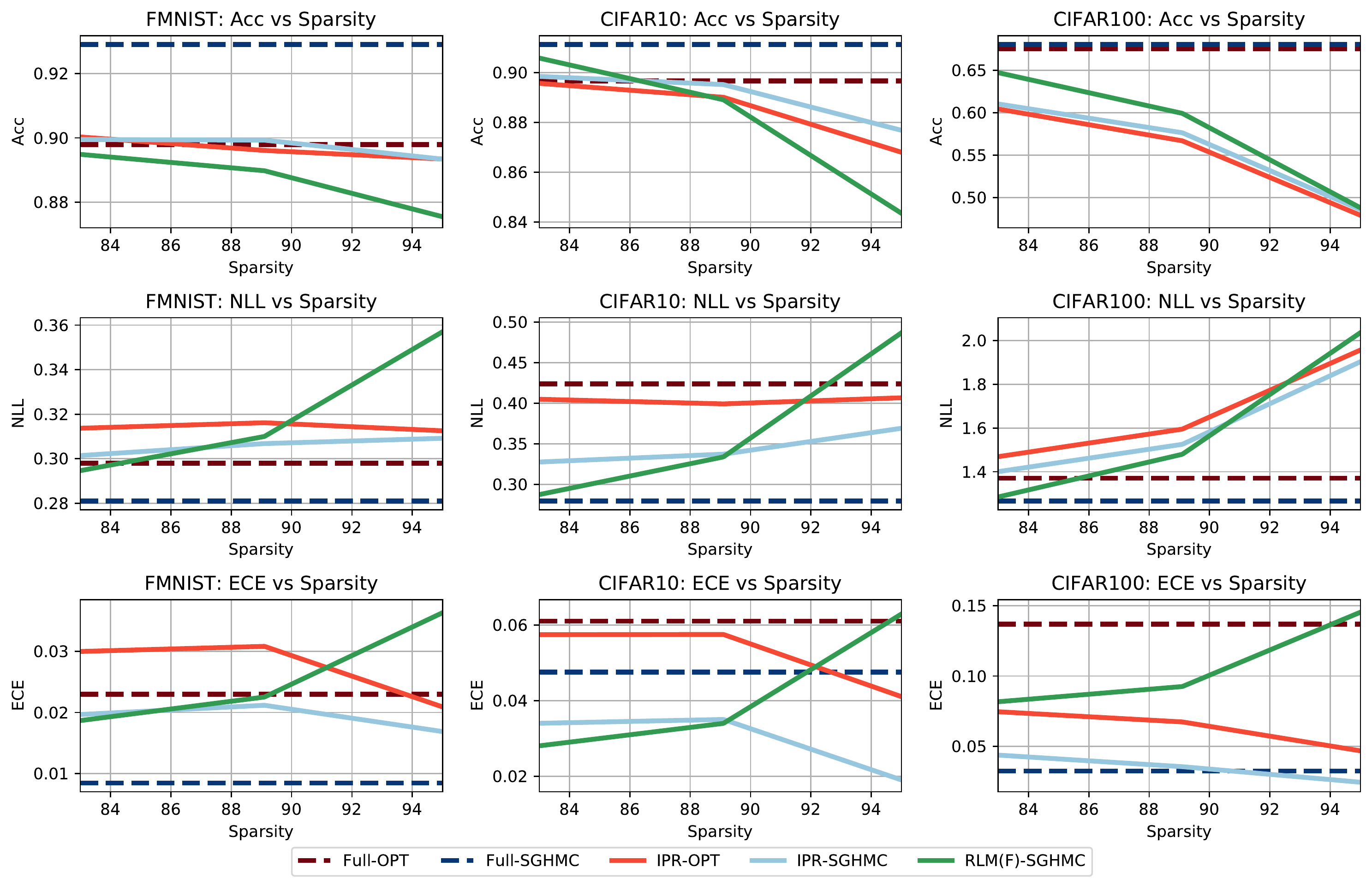}
\caption{Performance of SGHMC and optimization-based learning on the FMNIST, CIFAR10 and CIFAR100 datasets.}
\label{fig:app-exp4}
\end{figure*}

In this section we include additional results from the main paper. Figures \ref{fig:app-exp3-1} - \ref{fig:app-exp3-3} contain additional results from Experiment 2, examining the performance of SGHMC on random sparse sub-structures. Figures \ref{fig:ACF_95} \& \ref{fig:ACF_95_CIFAR100} contains additional ACF and iNLL from Experiment 3. Figure \ref{fig:app-exp4} contains the full panel of results from Experiment 4.

\begin{table}[htbp]
\small
\center
\caption{Time comparison of inference on a CPU.}
\begin{tabular}{llcccc}
\toprule
 Dataset &      Method & Sparsity &  \begin{tabular}[c]{@{}c@{}}Inference Time/\\ data point (s)\end{tabular} &  \begin{tabular}[c]{@{}c@{}}Inference\\speedup\end{tabular} & Hardware \\
\midrule
  FMNIST &  Full-SGHMC &       0\% &          0.0104 & $1\times$  & \multirow{4}{*}{\begin{tabular}[c]{@{}c@{}}Apple Macbook Pro\\ (8 GB memory, M1)\end{tabular}} \\
  FMNIST &    IP-SGHMC &      83\% &          0.0097 & $1.03\times$ & \\
  FMNIST &    IP-SGHMC &      89\% &          0.0067 & $1.55\times$ & \\
  FMNIST &    IP-SGHMC &      95\% &          0.0032 & $3.25\times$ &  \\
  FMNIST &    IP-SGHMC &      99.5\% &          0.0006 & $17.33\times$ &  \\
  FMNIST &    Full-OPT &       0\% &           0.0002 & $50\times$ & \\
\midrule
  FMNIST &  Full-SGHMC &       0\% &          0.0139 & $1\times$ & \multirow{5}{*}{\begin{tabular}[c]{@{}c@{}}Google Colab\\ (13 GB memory, Intel Xeon@2.20GHz)\end{tabular}} \\
  FMNIST &    IP-SGHMC &      83\% &          0.0104 & $1.33\times$ & \\
  FMNIST &    IP-SGHMC &      89\% &          0.0069 & $2\times$ & \\
  FMNIST &    IP-SGHMC &      95\% &           0.0031 & $4.48\times$ &  \\
  FMNIST &    IP-SGHMC &      99.5\% &           0.0007 & $19.86\times$ &  \\
  FMNIST &    Full-OPT &       0\% &           0.0003 & $50\times$ & \\
  \bottomrule
\end{tabular}

\label{tab:exp1_timing}
\end{table}

\begin{figure}[htbp]
\centering
\begin{subfigure}{.49\textwidth}
  \centering
  \includegraphics[width=\textwidth]{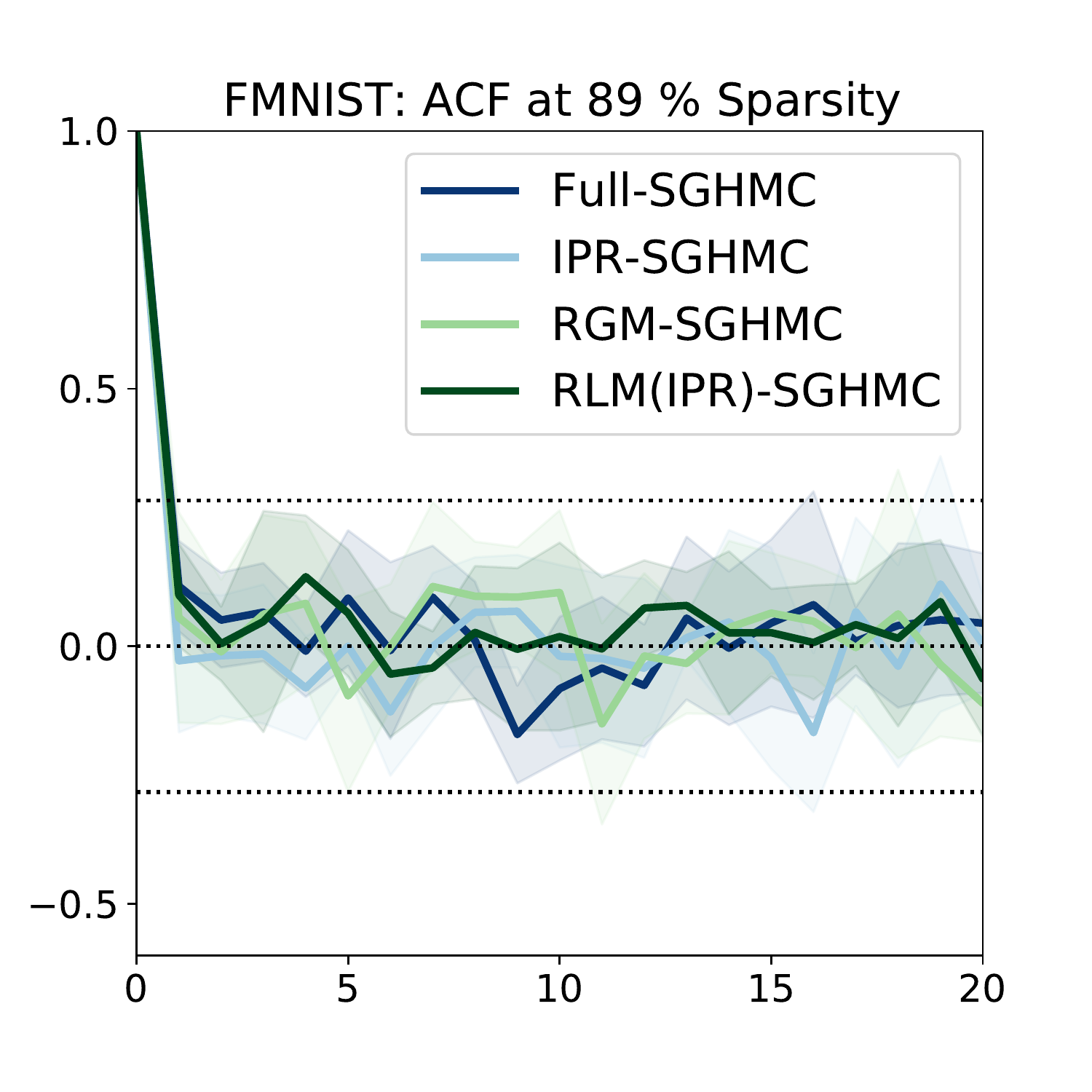}
\end{subfigure}
\begin{subfigure}{.49\textwidth}
  \centering
  \includegraphics[width=\textwidth]{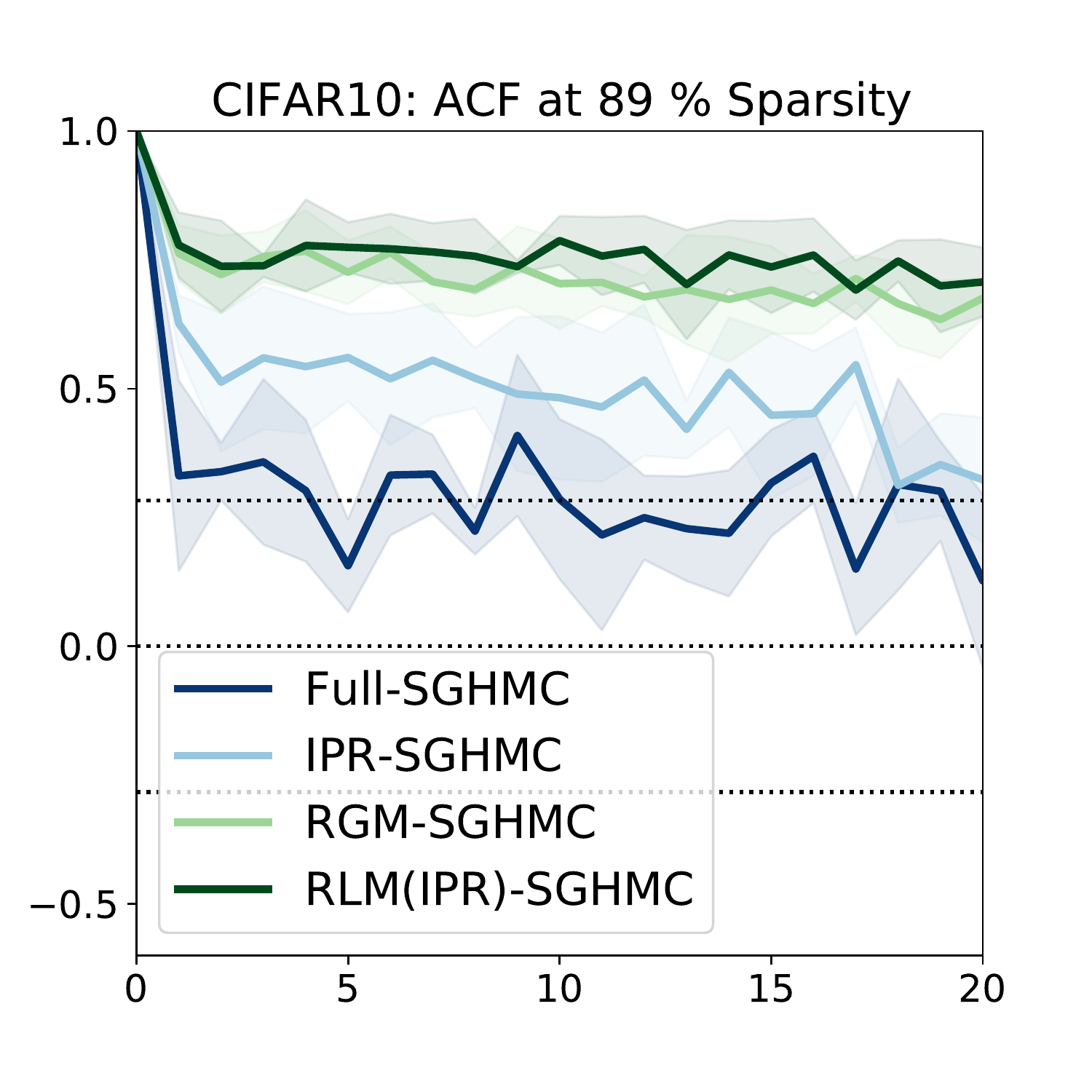}
\end{subfigure}
\caption{ACF plots for FMNIST and CIFAR10. The plots correspond to the case when sparsity level is set to 89\%. We also include the full model as a baseline.}
\label{fig:ACF_89}
\end{figure}

\begin{table}[htbp]
\small
\centering
\caption{\textbf{MLP200/FMNIST:} Mean effective sample size. The same random mask must be used for all chains here for a fair application of the ESS across the chains. This is over the 50 samples post burn-in.}
\label{tab:basic_results}
\begin{tabular}{lccc}
\toprule
Pruned &0 \% & 89.1 \%& 95.3 \% \\ \midrule
Full-SGHMC(5) & 3.1969 & 4.6081 & 5.7829\\
RLM(IPR)-SGHMC(5) & - & 2.6937 & 2.7511\\
RLM(F)-SGHMC(5) & - & 3.0512 & 3.6698\\
RGM-SGHMC(5) & - & 3.1178 & 3.5895\\
\bottomrule
\end{tabular}%
\end{table}

\end{document}